%% file: main.tex
\documentclass[11pt]{article}

\usepackage[final]{acl}

\usepackage{times}
\usepackage{latexsym}

\usepackage{microtype}
\usepackage{inconsolata}
\usepackage{graphicx}
\usepackage{subcaption}
\usepackage{booktabs} 

\usepackage{algorithm}
\usepackage{algorithmic}

\usepackage{spverbatim}

\usepackage{amsmath,amssymb}
\usepackage{tabularx}
\usepackage{multirow}
\usepackage{xr}
\usepackage{fancyvrb}
\usepackage{fvextra}    

\usepackage{url}
\usepackage{xcolor}
\usepackage{soul}
\usepackage{enumitem}
\sethlcolor{yellow}
\usepackage{float}
\usepackage{caption}
\usepackage{array,siunitx,adjustbox,makecell}

\usepackage{hyperref}

\usepackage{mathtools}
\usepackage{amsthm}

\usepackage{listings}
\usepackage[T1]{fontenc}
\usepackage[utf8]{inputenc}

\lstdefinestyle{pythonstyle}{
  breaklines=true,
  breakindent=0pt
}

\usepackage[capitalize,noabbrev]{cleveref}

\theoremstyle{plain}

\theoremstyle{definition}

\theoremstyle{remark}

\usepackage[textsize=tiny]{todonotes}

\title{InfoSynth: Information-Guided Benchmark Synthesis for LLMs}

\author{%
  Ishir Garg\\
  UC Berkeley\\
  {\tt ishirgarg@berkeley.edu}
  \And
  Neel Kolhe\\
  UC Berkeley\\
  {\tt neelkolhe@berkeley.edu}
  \AND
  Xuandong Zhao\\
  UC Berkeley\\
  {\tt xuandongzhao@berkeley.edu}
  \And
  Dawn Song\\
  UC Berkeley\\
  {\tt dawnsong@cs.berkeley.edu}
}

\begin{document}

\makeatletter
\let\@oldmaketitle\@maketitle
\renewcommand{\@maketitle}{\@oldmaketitle
  \begin{center}
    Project Page: \url{https://ishirgarg.github.io/infosynth_web/}
  \end{center}
  \vskip 0.2in
}
\makeatother

\maketitle

\begin{abstract}
Large language models (LLMs) have demonstrated significant advancements in reasoning and code generation, but efficiently creating new benchmarks to evaluate these capabilities remains a challenge. Traditional benchmark creation relies on manual human effort, which is expensive and time-consuming. Furthermore, existing benchmarks often contaminate LLM training data, necessitating novel and diverse benchmarks to accurately assess their genuine capabilities. This work introduces InfoSynth, a novel framework for automatically generating and evaluating reasoning benchmarks guided by information-theoretic principles. We propose metrics based on KL-divergence and entropy to quantify benchmark novelty and diversity without relying on costly model evaluations. Building on this framework, we develop an end-to-end pipeline that synthesizes robust Python coding problems from seed datasets using genetic algorithms and iterative code feedback. Our method generates accurate test cases and solutions to new problems 97\% of the time, and the synthesized benchmarks consistently exhibit higher difficulty compared to prior works. Moreover, our algorithm provides a method for controlling the novelty/diversity and difficulty of generated problems. InfoSynth offers a scalable, self-verifying pipeline for constructing high-quality, challenging coding benchmarks for LLMs.
\end{abstract}

\input{sections/1-introduction}

\input{sections/2-related}
\input{sections/3-properties}

\input{sections/4-pipeline}
\input{sections/5-results}
\input{sections/6-conclusion}

\bibliography{citations}

\newpage
\appendix

\input{sections/appendix}

\end{document}

%% file: sections/1-introduction.tex
\section{Introduction}
Large language models (LLMs) have demonstrated impressive capabilities in code generation and reasoning. However, rigorously evaluating these reasoning abilities remains a significant challenge. While substantial effort has been invested in creating robust math and coding benchmarks \citep{live-code-bench, big-code-bench, mbpp, human-eval, gsm8k, code-pref}, their development often demands considerable human labor or extensive computational resources for problem and solution validation. Some existing approaches utilize a judge LLM to generate and verify new problems \citep{genetic-instruct, crowd-select, unleash-scale-synth}. However, this method can yield erroneous benchmarks, as the judge LLM may not reliably solve the generated problems. This paper focuses on Python coding problems, whose solutions can be verified by executing them in a code environment. Our novel pipeline leverages this executability to ensure the robustness of the generated problems.

Beyond the challenge of ensuring robustness, state-of-the-art (SOTA) reasoning models often overfit to their training data, leading to poor performance on out-of-distribution problems \citep{math-perturb}. Furthermore, recent studies have revealed that LLM training data is frequently contaminated by existing evaluation benchmarks, which can artificially inflate reported performance \citep{contamination-llm, contamination-quiz, data-contamination}. For instance, \citet{gsm1k} show that LLMs experience accuracy drops of up to 8\% on their novel GSM1k dataset, despite its similarity in difficulty to the widely used GSM8k. This underscores the critical need for new, contamination-free reasoning benchmarks to genuinely assess the capabilities of different models.

To address these pressing issues, our work emphasizes two crucial benchmark properties: \textit{novelty} and \textit{diversity}. While this work does not directly address the task of creating contamination-free benchmarks, we provide an improved method of generating benchmarks that cover more diverse and novel coding tasks. A novel benchmark should comprise problems distinct from existing datasets, thereby preventing models from achieving high scores through mere memorization of previously seen examples. Conversely, a diverse benchmark should encompass a broad spectrum of dissimilar problems, enhancing its resilience against model overfitting and providing a more comprehensive evaluation. Clearly, robust, novel, and diverse benchmarks are essential for the reliable evaluation of LLM reasoning abilities. Our work seeks to answer two fundamental questions: (1) How can we effectively measure the novelty and diversity of benchmarks? (2) How can we efficiently generate benchmarks that possess these desirable properties while ensuring their correctness and robustness?

Our main contributions can be summarized as: 
\begin{itemize}[leftmargin=*, nosep]
    \item We introduce an information-theoretic framework to quantify and compare the \textit{novelty} and \textit{diversity} of benchmarks, offering a principled approach to benchmark assessment without reliance on model evaluations.
    \item We propose and validate an end-to-end pipeline, InfoSynth, for efficiently synthesizing novel, diverse, and verifiably correct Python coding problems from seed datasets with genetic algorithms and iterative code feedback.
    \item Through extensive experiments, we demonstrate that \textbf{InfoSynth exhibits superior robustness and difficulty} compared to existing data generation methods. Our pipeline provides a method for increasing the novelty and diversity of generated problems and controlling their difficulty.
\end{itemize}

%% file: sections/2-related.tex
\section{Related Work}

\textbf{Synthetic Problem Generation.}
Previous work has explored generating novel synthetic datasets from high-quality seed benchmarks. \citet{self-instruct, genetic-instruct, promptcot, wizard-lm, opencodeinstruct} show that LLMs can generate new instructions from existing ones. \citet{kodcode, dstc, rltf, codet, acecoder, autocodebench} successfully used LLMs to generate unit tests for solution verification. Our end-to-end pipeline extends existing methods with code-execution environments to ensure robustness, novelty, and diversity.

\textbf{Benchmark Quality Assessment.}
Efficient, concrete analysis of benchmark quality remains an open problem. Prior work defines metrics for novelty, separability, and difficulty via test-taker performance \citep{auto-bencher, crowd-select}, proposes adaptive selection of novel problems to reduce evaluation cost \citep{reliable-eff-amortized-eval, xie2025agentsynth}, and develops similarity and difficulty scores for coding tasks \citep{prog-task-difficulty}. A key limitation is reliance on SOTA model performance, making these methods accurate but computationally expensive. Our approach analyzes novelty and diversity without requiring costly model evaluations.

%% file: sections/3-properties.tex
\section{Desirable Benchmark Properties}
We propose a framework for characterizing the novelty and diversity of benchmarks. Our new novelty metric uses the KL-divergence to capture how different the benchmark is from existing datasets, with the broader goal of creating benchmarks that are contamination-free. Similarly, our proposed diversity metric uses Shannon entropy to capture how much variety exists among the problems.

\subsection{An Information-Theory Based Framework for Benchmark Analysis}
Formally, a baseline dataset can be modeled as samples $X=\{x_i\} \subseteq \mathbb{R}^d$ drawn from some true distribution $p(x)$, and the new dataset that we want to compare against the baseline can be modeled as samples $Y=\{y_i\}\subseteq \mathbb{R}^d$ drawn from a distribution $q(x)$. Here, $x_i, y_i$ represent the embedding vectors of the problem statements in an embedding space $\mathbb{R}^d$. We define the \textbf{novelty} of the new dataset $Y$ to be the KL-divergence between the distributions
\begin{align}
\text{Novelty}(Y|X) &=
D_{KL}\bigl(q \,\|\, p\bigr) \notag \\
&=
\int_{\mathbb{R}^{d}}
q(\mathbf{x})
\log\frac{q(\mathbf{x})}{p(\mathbf{x})}
\,d\mathbf{x}
\end{align}
Note that we take the KL-divergence of $p$ with $q$ as the reference distribution to reward datasets where $q(x)$ is large and $p(x)$ is small, indicating that the dataset contains problems not in the distribution of the seed dataset. Given a dataset $X = \{x_i\}, x_i \sim p(x)$, we define the \textbf{diversity} of the dataset to be the differential entropy of its distribution.
\begin{align}
\text{Diversity}(X) = -\int_{\mathbb{R}^{d}}
p(\mathbf{x})
\log p(\mathbf{x})
\,d\mathbf{x}
\end{align}
Intuitively, the KL-divergence captures the fact that novel datasets should have different embeddings from existing datasets. Similarly, diverse datasets should have embeddings that are fairly spread out, as clusters indicate problems that are likely to be similar and not diverse. A ``perfectly diverse'' dataset should resemble a uniform distribution over the embedding space so that it covers a large class of problems. Since the uniform distribution maximizes entropy, our metric captures this intuitive characterization of diversity.

In practice, obtaining the full distribution of the embedding space is intractable. Instead, we use statistical estimators for the KL-divergence and differential entropy.  
Given samples $x_1, ..., x_n, y_1, ..., y_m \in \mathbb{R}^d$ where $x$ and $y$ are drawn from $p(x), q(x)$ respectively, we use the k-NN based estimator by \citet{kl-div}
\begin{equation}\label{eq:kl_div_estimator}
D_{KL}(q || p) = \frac{d}{m}\sum_{i=1}^m\log\frac{\nu_k(i)}{\rho_k(i)} + \log\frac{n}{m-1}
\end{equation}
where $\nu_k(i)$ is the distance from $y_i$ to its $k$-th nearest neighbor in $\{x_j\}$ and $\rho_k(i)$ is the distance from $y_i$ to its $k$-th nearest neighbor in $\{y_j \mid j \neq i\};$ $k$ is a hyperparameter. Similarly, we can estimate the entropy of a dataset. Given samples $x_1, ..., x_N \in \mathbb{R}^d$, the Kozachenko-Leonenko estimator is
\begin{multline}\label{eq:entropy_estimator}
h(X) = \psi(N) - \psi(k) + \log V_d \\ + \frac{d}{N}\sum_{i=1}^N \log \rho_k(i)
\end{multline}
where $\psi$ is the digamma function, $V_d$ is the volume of the unit ball in $\mathbb{R}^d$, and $\rho_k(i)$ is the distance between $x_i$ and its $k$-th nearest neighbor in $\{x_j \mid j \neq i\}$; $k$ is a hyperparameter.

Computing embeddings, KL-divergence, and entropy for a text dataset is significantly faster and cheaper than computing test-taker statistics. Hence, we provide a way to cheaply estimate the quality of new benchmarks.

\subsection{Empirical Validation}
We empirically verify our metrics on existing datasets, embedding questions in $\mathbb{R}^{768}$ with all-mpnet-base-v2 \citep{song2020mpnet}. Since entropy estimation suffers from the curse of dimensionality, we project down via UMAP \citep{umap}; any datasets to be compared must share a single UMAP call to preserve their relative geometry. We renormalize after projection so distances correspond to cosine similarity.

\subsubsection{KL-divergence Metric Validation}

\begin{figure*}[ht]
    \centering
\includegraphics[height=0.13\textheight,keepaspectratio]{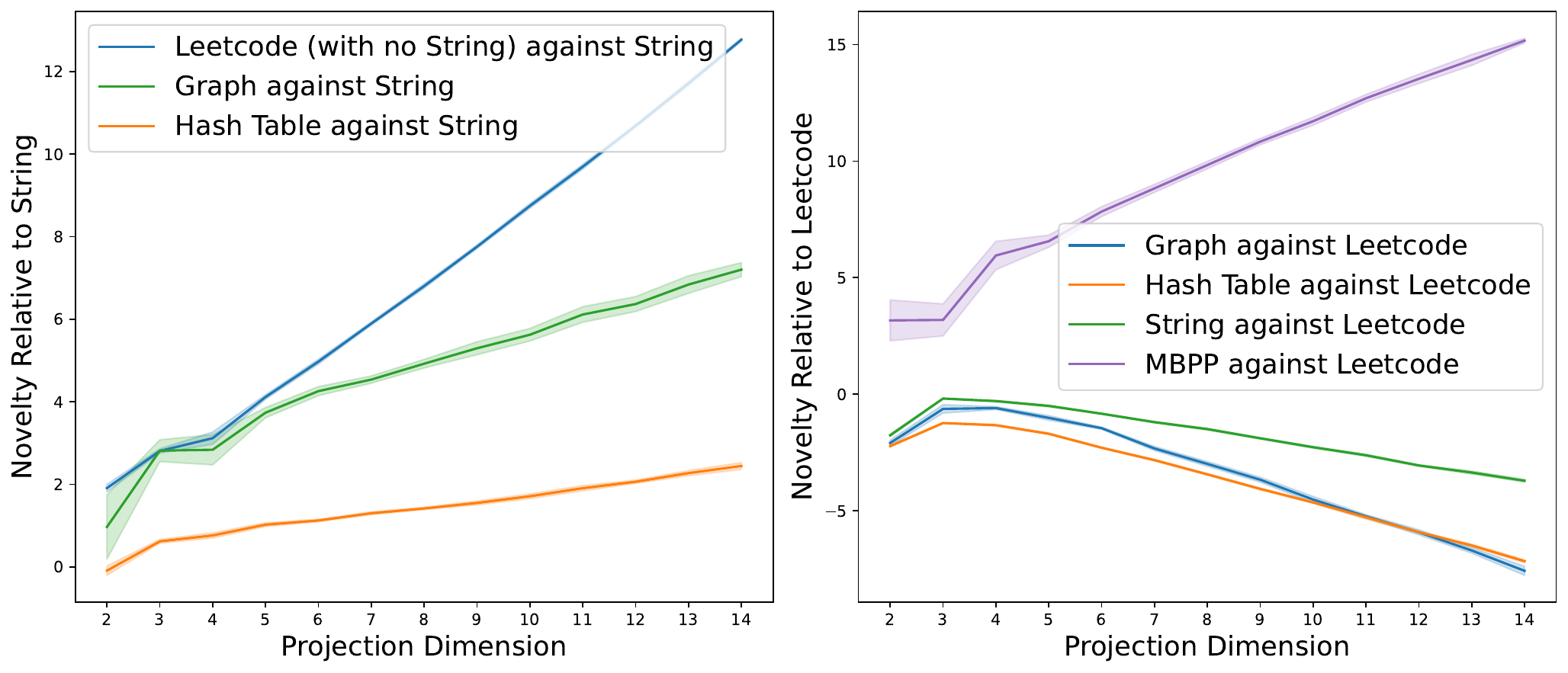}
\includegraphics[height=0.12\textheight,keepaspectratio]{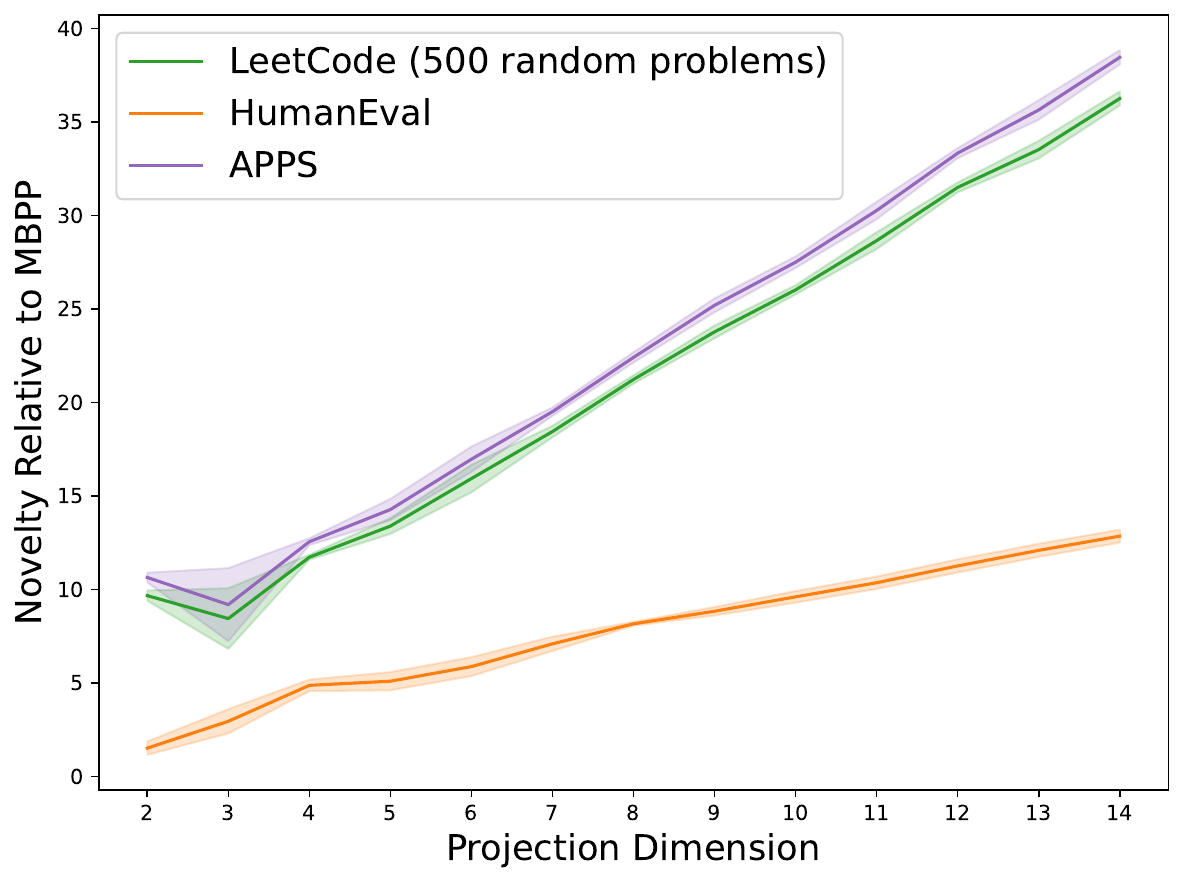}
    \caption{Left: Leetcode has higher novelty than its Hash Table and String subsets. Middle: MBPP has high novelty against Leetcode, while Leetcode subsets have low relative novelty. Right: Leetcode and APPS have high novelty against MBPP (harder, dissimilar problems); HumanEval is low (known to be MBPP-similar). 95\% CIs.}
    \label{fig:novelty_lc_string_validation}
\end{figure*}

We use 3511 Leetcode problems with per-problem concept labels \citep{leetcode-kl-div}, from which we extract three subsets by tag: ``Hash Map'' (686), ``Graph'' (160), and ``String'' (786). We also use the MBPP test set (374) \citep{mbpp}, HumanEval (164), and 500 randomly chosen APPS problems \citep{human-eval, apps-dataset}. UMAP uses 80 neighbors and min-distance 0.1, averaged over 10 independent runs with $k=4$ for the novelty estimator.

Figure~\ref{fig:novelty_lc_string_validation} shows our results align with intuition, confirming KL-divergence as a measure of benchmark novelty. The middle plot's estimator goes negative despite KL being theoretically nonnegative: comparing a subset against its superset yields small subset–superset distances but large intra-subset distances. In practice this never occurs in useful comparisons; we include it only to show that the metric still tracks human intuition, since we only care about \textit{relative} differences between datasets.

\subsubsection{Differential Entropy Metric Validation}
\begin{figure*}[htbp]
\centering

\includegraphics[height=0.185\textwidth,keepaspectratio]{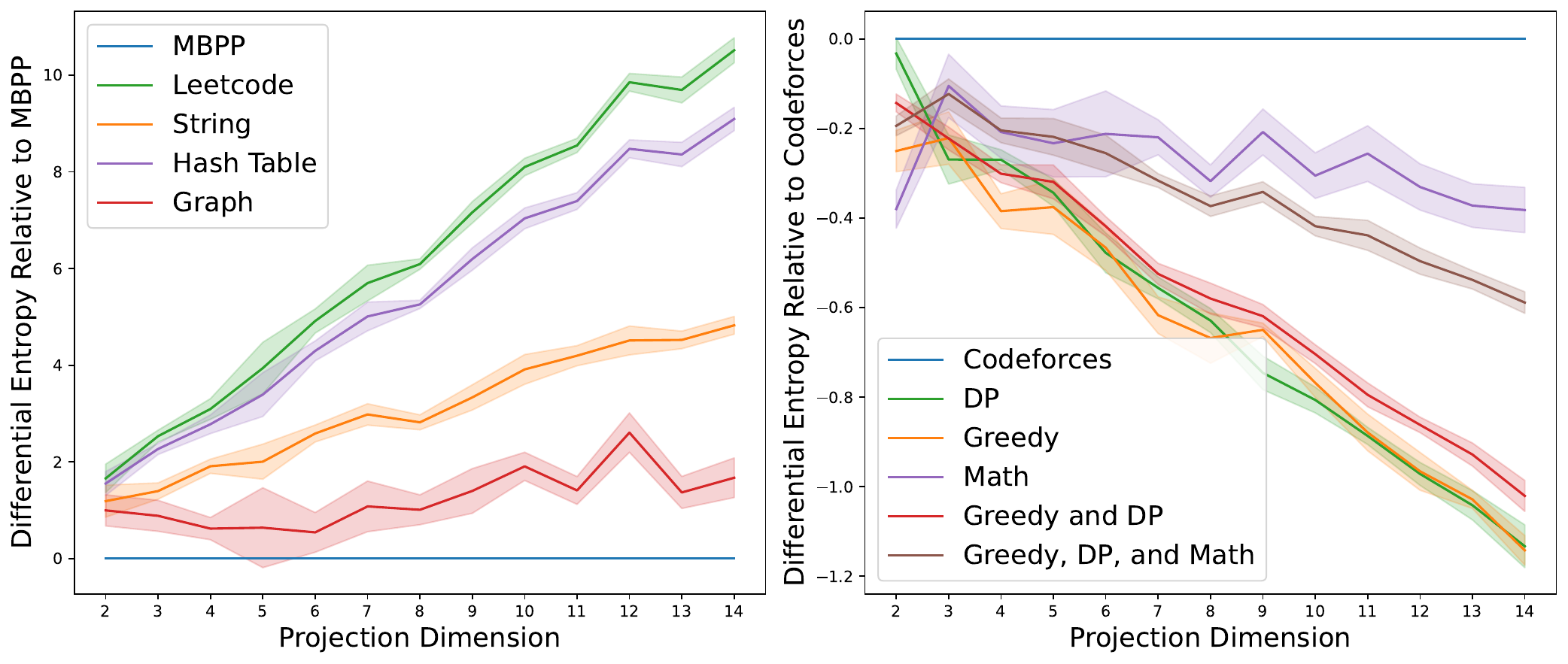}%
\includegraphics[height=0.185\textwidth,keepaspectratio]{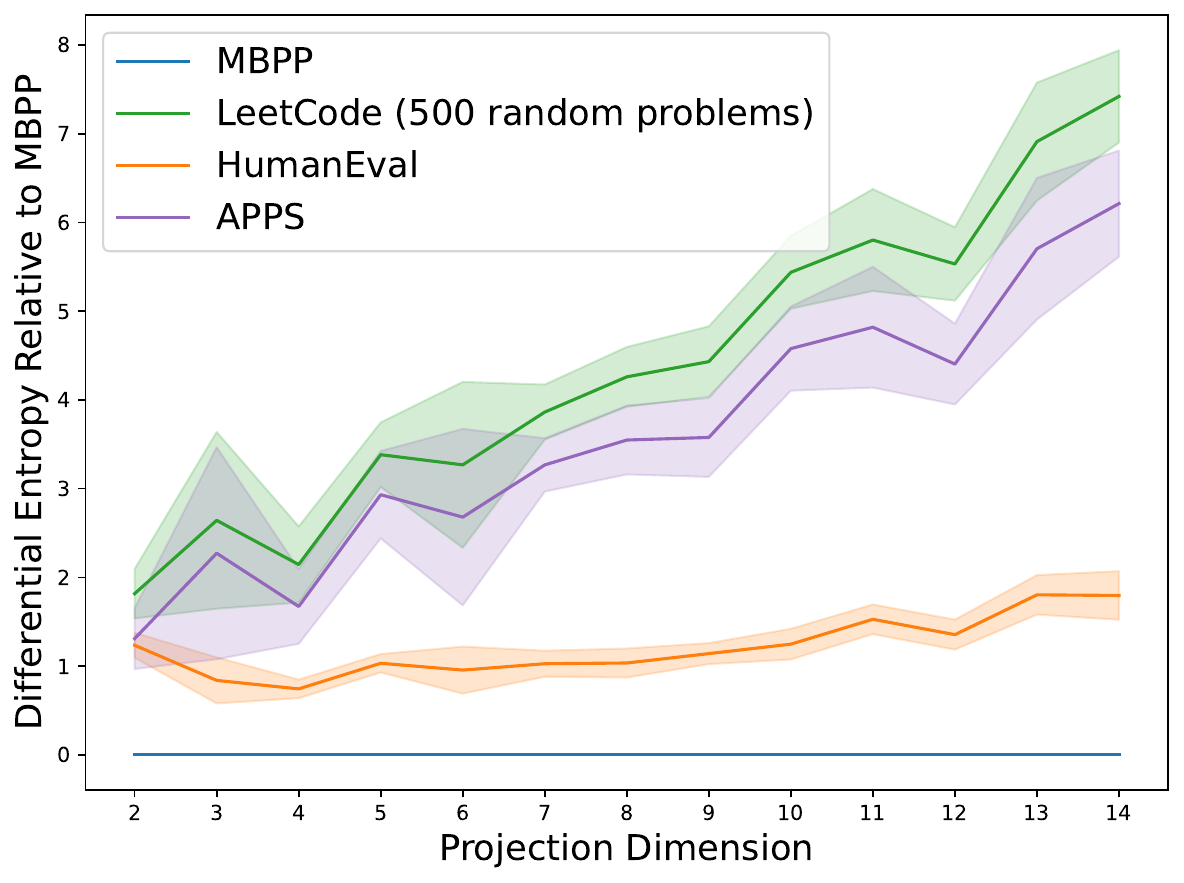}

\vspace{-3mm}

\caption{Left: Leetcode vs.\ MBPP entropy; MBPP is lower due to simpler, repetitive problems. Middle: Codeforces vs.\ subsets; full datasets exceed topic-specific subsets, except Math, which overlaps others (e.g., DP, Greedy) and appears highly dispersed when isolated. Right: Leetcode and APPS span many CS topics (high diversity); HumanEval is lower (easier, MBPP-similar). 95\% CIs.}
\label{fig:codeforces_lc_diversity_validation}
\end{figure*}

We add 4,000 random Codeforces problems \citep{openr1_codeforces} to the previous datasets, running UMAP with 80 neighbors and min-distance 0.1 over 10 trials. Because the Kozachenko--Leonenko estimator's k-NN distances shrink with dataset size, larger datasets yield artificially lower entropy; we therefore sample $N$ points per dataset without replacement many times and take the average.

We use $N{=}150$ over 250 trials ($k{=}4$) for Leetcode vs.\ MBPP and $N{=}800$ over 250 trials ($k{=}21$) for Codeforces. ``Diversity relative to X'' is plotted as a difference for visualization only; diversity itself is unary. Figure~\ref{fig:codeforces_lc_diversity_validation} shows that datasets expected to be more diverse exhibit higher entropy.

\subsection{Choosing \texorpdfstring{$k$}{k} and \texorpdfstring{$d$}{d}}
\citet{kl-k-rec} show that the bias–variance tradeoff for dataset size $N$ depends on $k/N$: larger $k$ increases bias but reduces variance and captures global structure. For KL-divergence, we use $k\!\approx\!4$ to emphasize local differences; for entropy, larger $k$ better captures global diversity, especially with scattered clusters; we find $k/N\!\in\![0.02,0.04]$ effective. Overall, diversity and entropy rankings are consistent across dimensions, and we recommend projecting to $d\!\in\![8,12]$ for dataset comparison. We present ablations for all hyper-parameters used to compute these metrics in Appendix \ref{app:sensitivity}.

%% file: sections/4-pipeline.tex
\section{A Novel Benchmark Synthesis Pipeline}
\label{pipeline-desc}
\begin{figure*}[t]
  \centering
    \resizebox{0.67\textwidth}{!}{\includegraphics{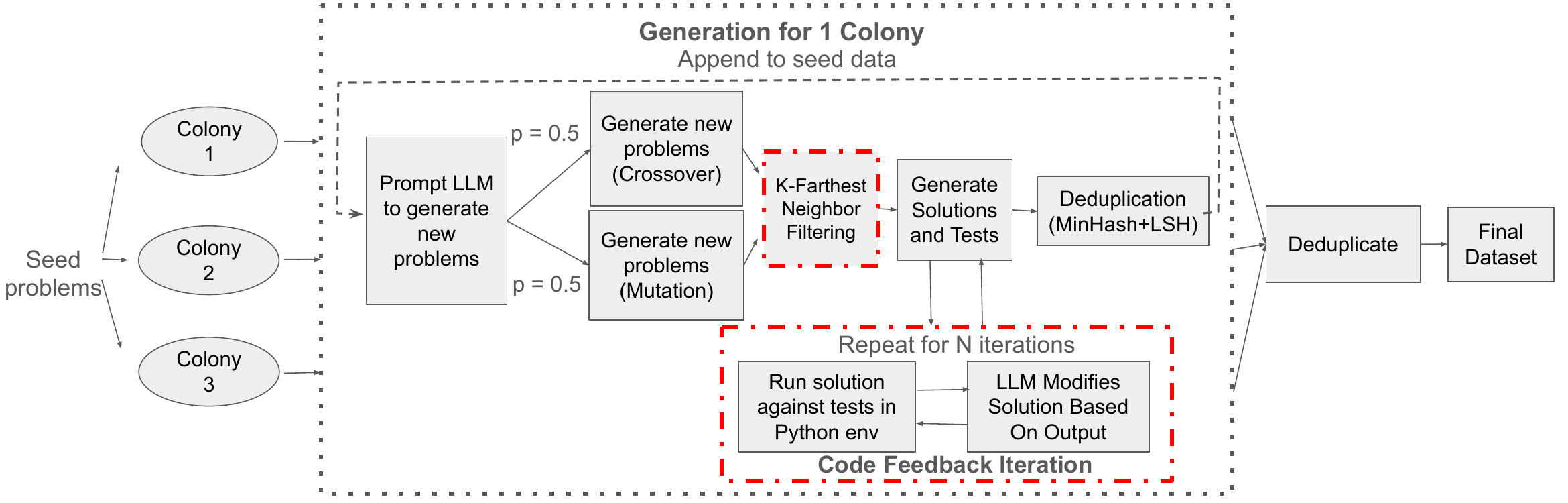}}
    \caption{Generation Pipeline. Each colony processes a subset of seeds via mutation or crossover per iteration, generates solutions/tests refined through iterative testing, then deduplicates similar problems within the colony. Survivors seed the next iteration; colony outputs are merged and deduplicated into the final dataset.}
  \label{fig:pipeline}
\end{figure*}

We introduce an end-to-end genetic algorithm for generating novel and diverse datasets from a seed dataset. The detailed algorithm is given in Appendix \ref{app:algo}, but we describe the main ideas here. Figure \ref{fig:pipeline} provides a visual outline of the pipeline.
\subsection{Data Generation Pipeline}
\textbf{Mutation and Crossover.}
We randomly apply either crossover or mutation to the existing data at each iteration to generate new coding instructions \citep{genetic-instruct}. A key change from previous work is that our mutation modifies an existing problem in three difficulty variations: easier, equally difficult, and harder, encouraging diversity. We demonstrate the benefit of this in Section \ref{effect_vary_diff}. Crossover prompts combine existing questions into new ones. The examples below show how mutation and crossover produce problems.
\par\medskip\noindent\makebox[\linewidth]{\fbox{\begin{minipage}{0.45\textwidth}
\textbf{Seed Question:} Write a Python function to find the sum of an array.
\medskip
\textbf{Hard Mutation Variant:} Write a Python function to find the sum of an array, where the array may contain nested lists of integers at any depth.
\end{minipage}}}
\par\medskip
\par\medskip\noindent\makebox[\linewidth]{\fbox{\begin{minipage}{0.45\textwidth}
\textbf{Seed Questions:} 1. Write a function to rotate a given list by a specified number of items to the right direction. 2. Write a function to find the maximum sum that can be formed which has no three consecutive elements present.
\medskip
\textbf{Crossover Variant:} Write a function to rotate a list by a specified number of steps to the right, ensuring that the sum of any three consecutive elements in the newly rotated list does not exceed a given threshold.
\end{minipage}}}
\par\medskip 
\noindent
Appendix \ref{app:problem-example} contains more examples of problems. Prompts are given in Appendix \ref{app:mutation}, \ref{app:crossover}.

\textbf{k-Farthest Neighbor Selection.}
A key improvement of InfoSynth is that in order to increase novelty and diversity, we filter problems by cosine similarity to those already generated. In mutation, we produce easy, medium, and hard variants, retaining the two of three with lowest similarity to the seed and generated set. In crossover, we likewise generate three problems and keep the two least similar to the dataset.

\textbf{Iterative Code Feedback.}
For each new problem, the model generates a Python solution and test cases (prompts in Appendix~\ref{app:solution-tests}, \ref{app:solution-tests-feedback}). Candidate solutions are executed in an isolated environment, and the results are fed back to the model, which iteratively refines its solution and tests until all tests pass or a maximum number of iterations is reached. A key improvement of InfoSynth over prior methods is feeding the entire feedback history at each step, giving the model richer context; Section~\ref{iterative-code-feedback} explains how this induces chain-of-thought reasoning \citep{cot}. Importantly, problems failing self-verification are excluded from the final dataset but still serve as seeds for the next generation round to encourage diversity.

\textbf{Deduplication.}
We use the MinHash + LSH algorithm with 250 permutations and a 0.75 similarity threshold to remove textually similar problems, similar to that done by \citet{genetic-instruct}.

\textbf{Postprocessing.}
Generated problem descriptions are not always well-aligned with their test cases. For example, a problem may not describe how to handle edge-cases such as null inputs or empty arrays. In some problems, it is unreasonable to expect a test-taker to infer the desired behavior (e.g., should we return None or -1 on an empty array input?). An example of such a problem is given in Appendix \ref{app:post-processed}. For each problem-test pair, the model is prompted to rephrase the question to incorporate details on handling obscure edge-cases. The prompt is given in Appendix \ref{app:postprocess-prompt}.

\subsection{Experimental Setup}
We generate six Python coding datasets in English using GPT-4o \citep{gpt4o-card} as the generator. The first dataset, called MBPP-Guided, is seeded with MBPP. The second dataset, MBPP-Hard-Guided, is also seeded with MBPP, but during mutation, the model is prompted to only make the questions more difficult. The third dataset, Leetcode-Guided, is seeded with a Leetcode dataset developed by \citet{leetcode-seed}. For each of these three datasets, we perform the generation process again, but without using k-farthest neighbor selection, resulting in a total of six datasets. These additional datasets are referred to as MBPP-New, MBPP-Hard, and Leetcode-New, respectively.

%% file: sections/5-results.tex
\section{Results and Analysis}
\label{results}
We categorize generated problems as: (1) \textbf{Passing} (solution passes all tests), (2) \textbf{Failing} (fails $\geq$ 1 test), (3) \textbf{Erroring} (syntax/runtime error), and (4) \textbf{Unparsable} (malformed, e.g., missing [solution]/[test] tags, more common for smaller models). Table~\ref{tab:dataset_stats} reports benchmark statistics: test cases equal the number of \texttt{assert} statements, and test coverage is the fraction of code lines executed when all tests are run. We also evaluate SOTA models on all datasets (Table~\ref{tab:data}); Qwen2.5 models use 4-bit quantization. Since MBPP contains vague/misleading problems \citep{mbpp}, we post-processed it for fairer comparison. Thus, Table~\ref{tab:data} focuses on post-processed results as these provide the fairest comparison, omitting filtered versions (see Section~\ref{postprocessing-step}). For each benchmark, we randomly sample 100 problems and manually verify that their solutions and test cases are fully correct and consistent with their problem statements.

\input{dataset_info}
\input{data}

\subsection{Novelty and Diversity Analysis}

\begin{figure}[!t]
\centering

\begin{subfigure}{1\columnwidth}
    \centering
    \includegraphics[width=\linewidth]{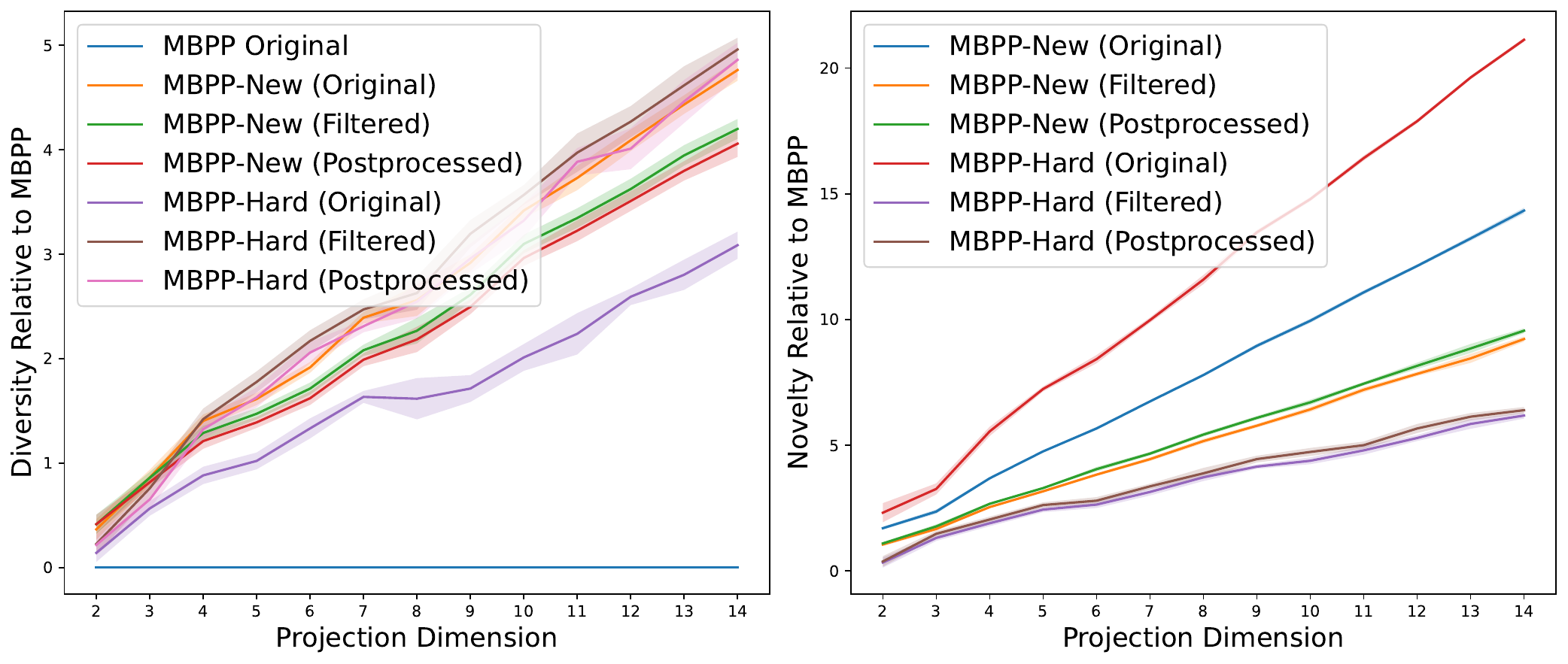}
    \caption{Novelty and diversity in MBPP-New and MBPP-Hard. Using all three mutation types boosts novelty. MBPP-Hard has higher diversity than MBPP-Original.}
    \label{fig:new_mbpp_novelty}
\end{subfigure}

\vspace{0em}

\begin{subfigure}{1\columnwidth}
    \centering
    \includegraphics[width=\linewidth]{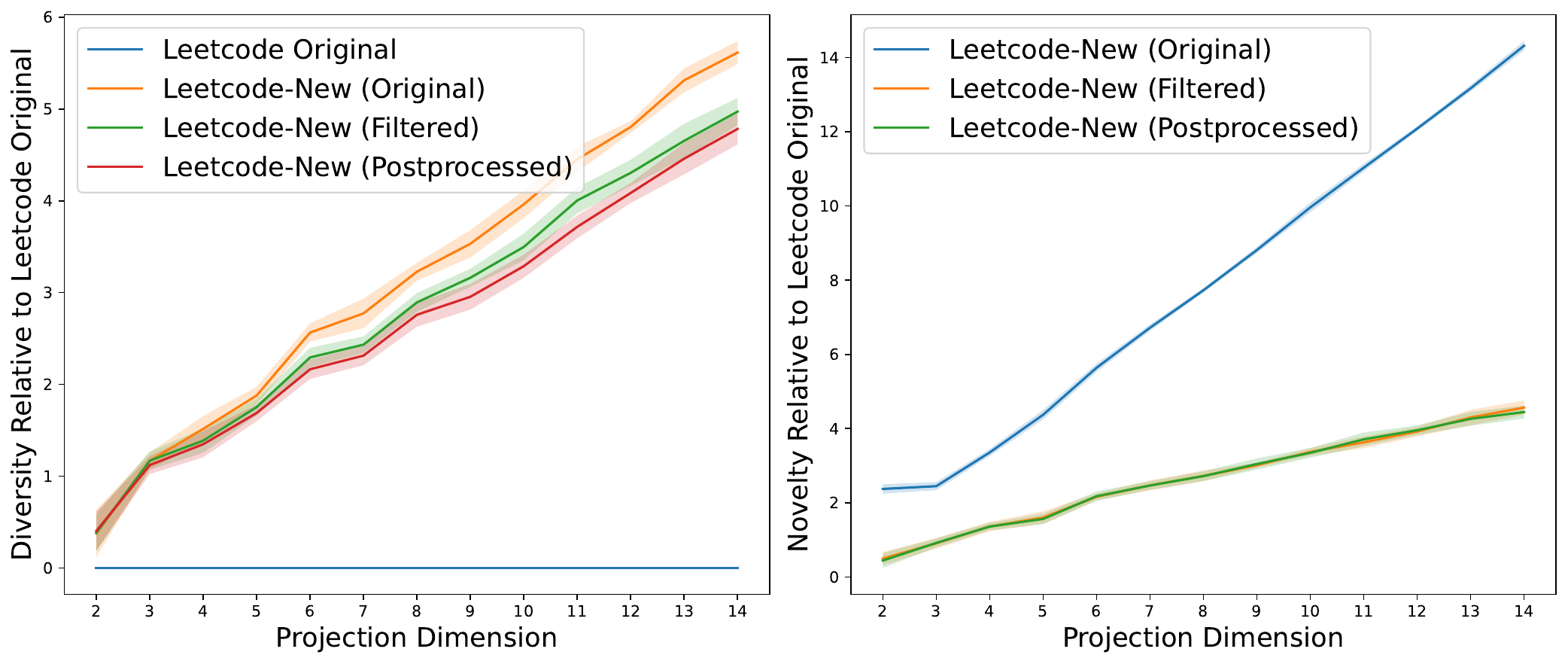}
    \caption{Leetcode-New shows greater diversity compared to the original Leetcode.}
    \label{fig:new_lc_novelty}
\end{subfigure}

\vspace{0em}

\begin{subfigure}{1\columnwidth}
    \centering
    \includegraphics[width=\linewidth]{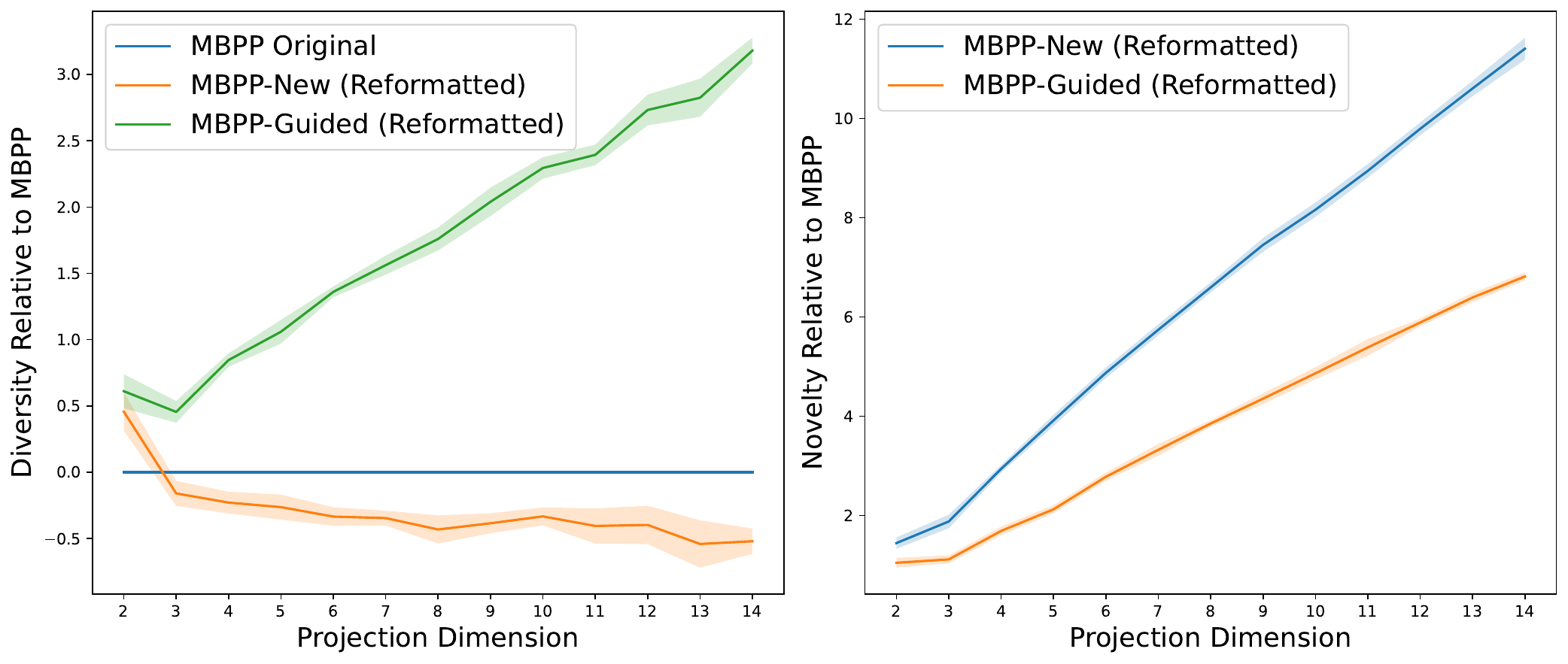}
    \caption{MBPP-Guided exhibits higher diversity but reduced novelty compared to MBPP-New.}
    \label{fig:mbpp_guided_novelty_diversity}
\end{subfigure}

\vspace{0em}

\begin{subfigure}{1\columnwidth}
    \centering
    \includegraphics[width=\linewidth]{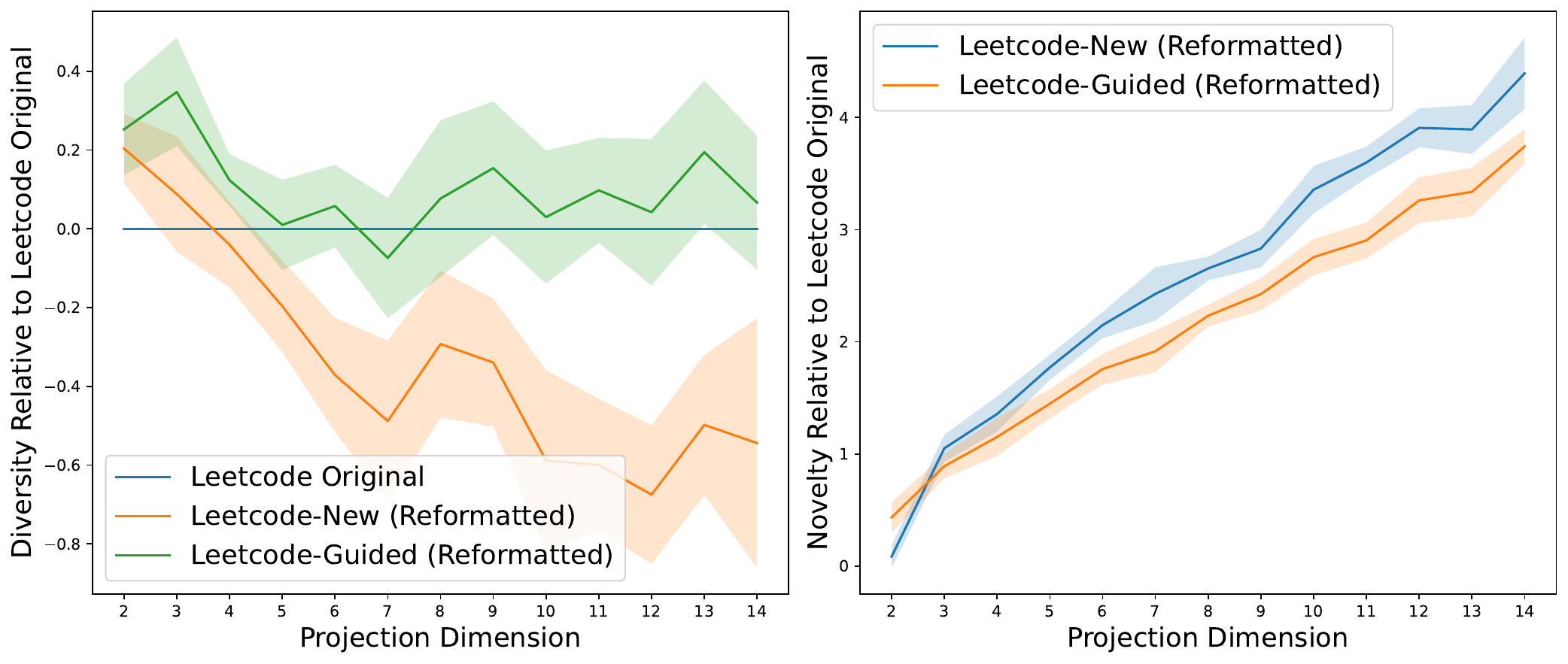}
    \caption{Leetcode-Guided exhibits higher diversity but reduced novelty compared to Leetcode-New.}
    \label{fig:lc_guided_novelty_diversity}
\end{subfigure}

\caption{Novelty and diversity analysis across MBPP and Leetcode variants.}
\label{fig:novelty_diversity_all}
\end{figure}

Figure \ref{fig:new_mbpp_novelty} presents the novelty and diversity of MBPP-New and MBPP-Hard relative to MBPP-Original, while Figure \ref{fig:new_lc_novelty} shows the same comparison for Leetcode-New relative to Leetcode-Original. Overall, our pipeline produces datasets that are more novel and diverse than the original seeds. However, filtering and post-processing reduce novelty compared to the initial generation. We attribute this to LLM memorization \citep{math-perturb, llm-memorization}, since more novel problems are often out-of-distribution and harder for the model to solve. All UMAP simulations use 80 neighbors and a minimum distance of 0.1, except for Figure \ref{fig:lc_guided_novelty_diversity}, which uses 30 neighbors due to smaller dataset size. In general, our results are not sensitive to UMAP hyperparameters.

Figures \ref{fig:mbpp_guided_novelty_diversity} and \ref{fig:lc_guided_novelty_diversity} show that k-farthest-neighbor filtering improves dataset novelty and diversity. This comes at the cost of generating easier problems (Table \ref{tab:data}), highlighting the ability of InfoSynth to control the novelty--diversity--difficulty tradeoff. Empirically, this arises because the generator struggles to produce difficult problems that are not conceptually aligned with the seeds. We also observe a tradeoff between novelty and diversity: highly novel datasets tend to concentrate around low-density regions in the seed-embedding distribution, which increases novelty but reduces diversity. Our results also show that filtering and post-processing reliably improve diversity.





\subsection{Effect of Varying Mutation Difficulties}
\label{effect_vary_diff}
 Table \ref{tab:data} shows MBPP-Hard scores are 8\%-15\% lower than MBPP-Original across most models, suggesting hard mutations effectively raise difficulty. This comes with a tradeoff of the dataset having reduced diversity and novelty as the problems tend to be concentrated around fewer, but more challenging topics. Hence, the set of mutation difficulties can be chosen to control the difficulty of the produced benchmark.

\begin{table*}[t]
\centering
\small
\setlength{\tabcolsep}{6pt}
\renewcommand{\arraystretch}{1}

\resizebox{\textwidth}{!}{%
\begin{tabular}{lccccc}
\toprule
\textbf{Model} & \textbf{InfoSynth (ours)} & \textbf{KodCode} & \textbf{GeneticInstruct} & \textbf{AutoCodeBench} & \textbf{OpenCodeInstruct} \\
\midrule
o4-mini           & \textbf{48.47\%} & 75.15\% & 61.00\% & 72.00\% & 63.85\% \\
gpt-4.1-mini      & \textbf{48.47\%} & 69.00\% & 59.5\%  & 64.44\% & 72.59\% \\
gemini-2.0-flash  & \textbf{49.08\%} & 55.50\% & 51.00\% & 63.57\% & 80.00\% \\
claude-3.7-sonnet  & \textbf{42.33\%} & 53.00\% & 56.00\% & 78.29\% & 76.92\% \\
\bottomrule
\end{tabular}%
}
\caption{Comparison of model pass rates across different synthetic instruction generation methods. Lower means harder problems.}
\label{tab:infosynth-comparison-pass-rate}
\end{table*}

\subsection{Effect of Iterative Code Feedback}
\label{iterative-code-feedback}
We find that passing solution-test pairs increase by 20\% over 5 feedback iterations, showing the effectiveness of code iteration in producing robust problems. Error rates drop as the LLM fixes syntax/runtime issues, though the unparsable rate rises slightly due to occasional formatting failures. Appendix~\ref{app:code-feedback} shows feedback curves. Three iterations are typically ideal; further iterations yield marginal gains not worth the extra inference cost. We also find that iterative feedback acts as chain-of-thought (CoT) reasoning \citep{cot}, as the model leverages the full feedback history to refine solutions/tests, lowering both error and failure rates. An example of this is in Appendix~\ref{app:cot}.

\subsection{Analysis of Filtered-Out Problems}
We include an analysis of the types of problems filtered out in InfoSynth and the insights it provides in Appendix \ref{app:filtered-analysis}. InfoSynth reveals that current models struggle when handling interacting constraints and maintaining global consistency, get stuck in superficial debugging loops focused on local fixes without revisiting higher-level structure, and fail to produce numerical test cases even when they derive correct symbolic solutions. 

\subsection{Effect of the Postprocessing Step}
\label{postprocessing-step}
Appendix \ref{app:post-processed} shows two post-processed examples, where the model resolves ambiguous edge cases and sometimes rephrases statements more concisely without losing information. Table \ref{tab:data} shows 5--15\% accuracy gains across most test-taker models, confirming that post-processing reduces ambiguity. Manual verification of 100 problems per dataset further shows 100\% of post-processed problems are correctly reformatted without altering the core question.

\subsection{Relating Diversity and Topic Coverage}
For each problem in MBPP-Original, MBPP-New, and MBPP-Guided, we prompted GPT-4o-mini \citep{gpt4o-card} to list up to 3 topics describing the problem mimicking \citet{promptcot}. The list of allowed topics was taken from the Leetcode dataset \citep{leetcode-kl-div}. The prompt is given in Appendix \ref{topic-labeling}.

\begin{figure}[htbp]
\centering
\resizebox{1\columnwidth}{!}{\includegraphics{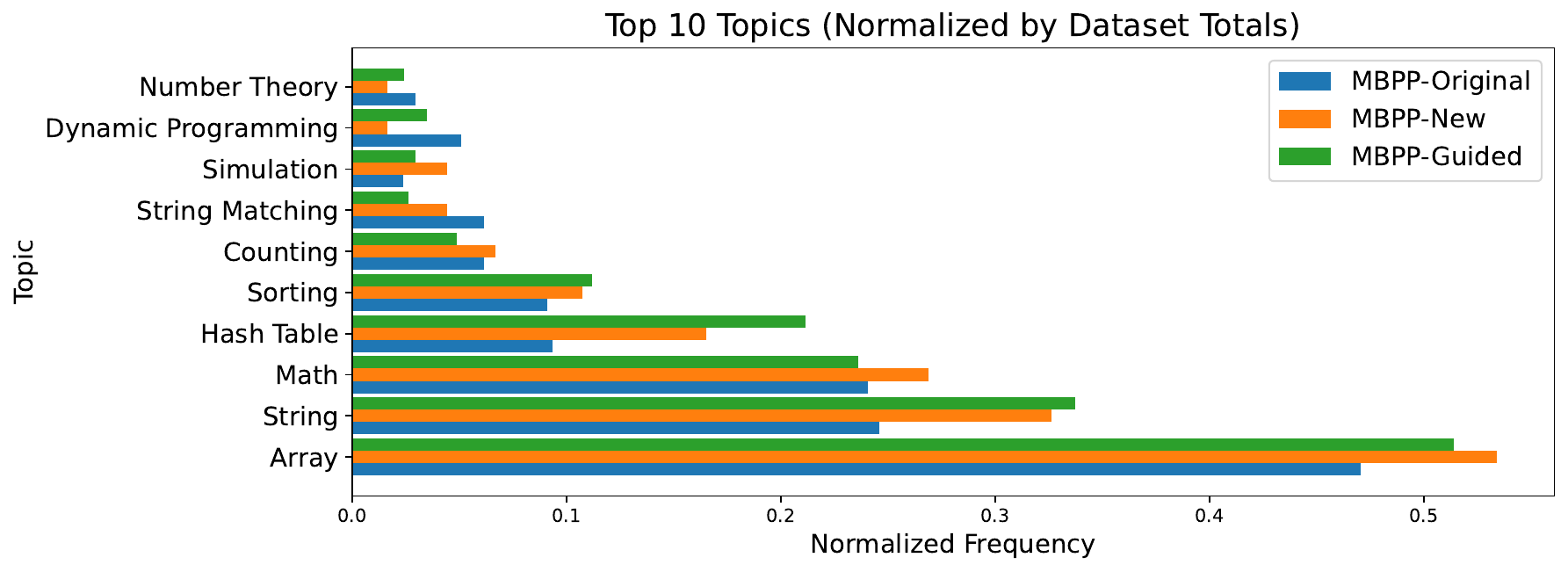}}
\caption{Fraction of problems relating to each topic for the 10 most common topics}
\label{fig:mbpp_topics}
\end{figure}

Figure \ref{fig:mbpp_topics} shows that InfoSynth increases the number of problems that use each concept for most topics, creating more diverse problems. We find that for topics with lesser coverage in the original MBPP dataset, our pipeline produces many more problems covering those topics.

\subsection{Comparison to Previous Methods}
We compare InfoSynth to GeneticInstruct \cite{genetic-instruct}, KodCode \cite{kodcode}, OpenCodeInstruct \cite{opencodeinstruct}, and AutoCodeBench \cite{autocodebench}. Note that all of pipelines have used Leetcode-based seed datasets.

\paragraph{Difficulty Comparison.} Table \ref{tab:infosynth-comparison-pass-rate} shows InfoSynth generates more challenging problems than prior works. The high pass rates on AutoCodeBench and OpenCodeInstruct partly reflect quality issues: a large fraction of OpenCodeInstruct lacks correct solutions/tests, so we filter using its provided correctness labels. This itself highlights an InfoSynth advantage: although OpenCodeInstruct uses a simpler mutation/crossover, it omits iterative feedback, which we find essential for raising difficulty. AutoCodeBench targets multi-language coverage rather than difficulty, applying only a basic filter that removes problems solvable 10 times in a row by a test-taker; InfoSynth instead combines mutation/crossover, iterative feedback, and novelty-based filtering.

\paragraph{Token Efficiency Comparison.}
Let $N$ denote the token length of a full problem instance (statement, solution, and tests). Table~\ref{tab:efficiency} summarizes per-instance token costs and their sources. InfoSynth uses iterative, execution-driven refinement: each problem runs multiple feedback rounds where the model generates a solution, executes it in a sandbox, and refines on errors ($\approx 7N$ input). KodCode samples $K$ independent solution attempts without feedback, yielding lower input but similar output cost ($\approx 4N$ across attempts). GeneticInstruct uses a fixed four-step pipeline (generate, solve, LLM-judge, deduplicate) with $\approx 4N$ input, relying on LLM-based evaluation rather than execution. AutoCodeBench adds four generation stages (problem, solution, test integration, quality control) plus up to 10 test-taker solve attempts per problem, scaling to $\approx 14N$. OpenCodeInstruct mirrors GeneticInstruct's generate--solve--judge--deduplicate pipeline but with extra judging/deduplication overhead, pushing output beyond $2N$. Overall, InfoSynth uses more cheap input tokens but a comparable count of expensive output tokens to prior work.

\begin{table}[htbp]
\centering
\small
\resizebox{0.5\textwidth}{!}{%
\begin{tabular}{lcc}
\hline
\textbf{Method} & \textbf{Input Tokens} & \textbf{Output Tokens} \\
\hline
InfoSynth & $7N$ & $4N$ \\
KodCode & $4N$ & $4N$ \\
GeneticInstruct & $4N$ & $N$ + judge/dedup \\
AutoCodeBench & $14N$ & $14N$ \\
OpenCodeInstruct & $4N$ & $2N$ + judge/dedup \\
\hline
\end{tabular}
}
\caption{Per-instance token cost comparison of benchmark generation pipelines.}
\label{tab:efficiency}
\end{table}

\subsection{Choosing an embedding model}
We test various embedding models on our datasets in Appendix \ref{app:emb-model}. Across models, the relative novelty and diversity between datasets remains the same despite some fluctuations in their magnitudes.

%% file: dataset_info.tex
\renewcommand{\arraystretch}{0.8}
\setlength{\aboverulesep}{0.5pt}
\setlength{\belowrulesep}{0.5pt}
\begin{table*}[!htbp]
  \centering
  \caption{Dataset statistics and quality measures. Gen. Size: Initial generation size; Filtered: Problems removed via filtering; Avg. Tests: Avg.\ \# test cases per problem; Human Correct: Human-verified correctness (\%); Coverage: Test coverage (\%); Hours: Person-hours spent generating.} \label{tab:dataset_stats}
  \resizebox{0.95\linewidth}{!}{
\begin{tabular}{lcccccc}
    \toprule
    \textbf{Dataset} 
      & \textbf{Gen. Size} 
      & \textbf{\# Filtered} 
      & \textbf{Avg. Tests} 
      & \textbf{\% Human Correct} 
      & \textbf{\% Coverage} 
      & \textbf{Hours} \\
    \midrule
    MBPP-New            & 1002 & 539 & 8.30 & 97\% & 99\% & 13\\
    MBPP-Guided         & 992 & 570 & 8.86 & 98\% & 99\% & 14 \\
    MBPP-Hard          & 1007 & 219 & 10.35 & 96\% & 100\% & 14 \\
    MBPP-Hard-Guided    & 994 & 468 & 8.86 & 96\% & 100\% & 15 \\
    Leetcode-New        & 997 & 163 & 8.22 & 98\% & 99\% & 25 \\
    Leetcode-Guided     & 991 & 177 & 8.66 & 97\% & 100\% & 27 \\
    \bottomrule
  \end{tabular}
  }
\end{table*}

%% file: data.tex
\renewcommand{\arraystretch}{0.8}
\setlength{\aboverulesep}{1pt}
\setlength{\belowrulesep}{1pt}
\begin{table*}[!htbp]
  \centering
    \caption{Test-taker performance on datasets}  \label{tab:data}
  \resizebox{0.9\linewidth}{!}{
  \begin{tabularx}{\textwidth}{l l *{3}{>{\centering\arraybackslash}X} *{3}{>{\centering\arraybackslash}X}}
    \toprule
    \multirow{2}{*}{Model}
      & \multirow{2}{*}{Dataset}
      & \multicolumn{3}{c}{Filtered}
      & \multicolumn{3}{c}{Postprocessed} \\
    \cmidrule(lr){3-5} \cmidrule(lr){6-8}
      & 
      & \%Pass & \%Fail & \%Err
      & \%Pass & \%Fail & \%Err \\
    \midrule

    \multirow{5}{*}{Qwen2.5-7b-Instruct}
      & MBPP-Original    & 52.67 & 45.99 & 1.34  & 60.43     & 37.70     & 1.87     \\
      & MBPP-New             & 34.32 & 61.60 & 4.08 & 45.83 & 51.58 & 2.60  \\
      & MBPP-Guided       & - & - & - & 54.04     & 44.39     & 1.58     \\
      & MBPP-Hard       & 19.82 & 72.52 & 7.66 & 30.14     & 63.01     & 6.85     \\
      & MBPP-Hard-Guided       & - & - & - & 39.10     & 56.84     & 4.06     \\
      & Leetcode-Original & -    & -     & -  & 11.84 & 83.77 & 4.39   \\
      & Leetcode-New         & 25.29 & 74.12 & 0.59 & 22.70 & 75.46 & 1.84  \\
      & Leetcode-Guided       & - & - & - & 28.25     & 71.19     & 0.56     \\
    \midrule

    \multirow{5}{*}{Qwen2.5-3b-Coder}
      & MBPP-Original    & 46.79 & 48.93 & 4.28 & 52.41     & 42.51     & 5.08      \\
      & MBPP-New             & 31.35 & 55.47 & 13.17 & 38.78 & 49.54 & 11.69 \\
      & MBPP-Guided       & - & - & - & 40.53     & 49.30     & 10.18     \\
      & MBPP-Hard       & 21.62 & 54.95 & 23.42 & 24.20     & 53.88     & 21.92     \\
      & MBPP-Hard-Guided       & - & - & - & 29.06     & 58.12     & 12.82     \\
      & Leetcode-Original & -    & -     & - & 2.63 & 85.53 & 11.84     \\
      & Leetcode-New         & 18.24 & 62.94 & 18.82& 20.25 & 66.87 & 12.88 \\
      & Leetcode-Guided       & - & - & - & 14.12     & 76.27     & 9.60     \\
    \midrule

    \multirow{5}{*}{GPT-4.1-Mini}
      & MBPP-Original     & 58.02 & 36.10 & 5.88  & 66.04     & 30.48     & 0.00    \\
      & MBPP-New             & 56.96 & 40.63 & 2.41 & 67.35 & 29.68 & 2.97  \\
      & MBPP-Guided       & - & - & - & 71.93     & 22.46     & 5.61     \\
      & MBPP-Hard       & 44.14 & 50.45 & 5.41 & 55.25 & 38.81 & 5.94  \\
      & MBPP-Hard-Guided       & - & - & - & 68.38     & 27.78     & 3.85     \\
      & Leetcode-Original & -    & -     & -  & 32.89 & 54.82 & 12.28   \\
      & Leetcode-New         & 45.28 & 38.24 & 16.47& 48.47 & 42.33 & 9.20  \\
      & Leetcode-Guided       & - & - & - & 48.02     & 46.89     & 5.08     \\
    \midrule

    \multirow{5}{*}{Gemini-2.0-Flash}
      & MBPP-Original      & 64.97 & 35.03 & 0.00   & 68.72     & 31.02     & 0.26   \\
      & MBPP-New             & 53.99 & 45.64 & 0.37 & 63.64 & 36.18 & 0.19  \\
      & MBPP-Guided       & - & - & - & 71.93     & 27.02     & 1.05     \\
      & MBPP-Hard       & 44.59 & 52.25 & 3.15 & 49.32 & 45.66 & 5.02  \\
      & MBPP-Hard-Guided       & - & - & - & 62.39     & 35.04     & 2.56     \\
      & Leetcode-Original & -    & -     & -    & 32.46 & 64.47 & 3.07   \\
      & Leetcode-New         & 44.71 & 54.12 & 1.18 & 49.08 & 48.47 & 2.45  \\
      & Leetcode-Guided       & - & - & - & 45.76     & 50.85     & 3.39     \\
    \midrule

    \multirow{5}{*}{Claude 3.7 Sonnet}
      & MBPP-Original      & 63.37 & 36.63 & 0.00 & 70.86     & 29.14    & 0.00    \\
      & MBPP-New             & 55.29 & 44.71 & 0.00 & 64.75 & 35.25 & 0.00  \\
      & MBPP-Guided       & - & - & - & 74.74     & 24.91     & 0.35     \\
      & MBPP-Hard       & 45.05 & 52.25 & 2.70 & 56.62 & 40.64 & 2.74  \\
      & MBPP-Hard-Guided       & - & - & - & 66.03     & 33.33     & 0.64     \\
      & Leetcode-Original & -    & -     & -  & 31.14 & 67.11 & 1.75    \\
      & Leetcode-New         & 44.71 & 54.71 & 0.59 & 42.33 & 57.67 & 0.00  \\
      & Leetcode-Guided       & - & - & - & 48.02     & 51.98     & 0.00     \\
    \midrule

    \multirow{5}{*}{o4-mini}
      & MBPP-Original      & 66.58 & 33.42 & 0.00 & 70.05     & 29.68     & 0.27     \\
      & MBPP-New             & 58.26 & 41.19 & 0.56 & 70.13 & 29.68 & 0.19  \\
      & MBPP-Guided       & - & - & - & 77.19     & 22.63     & 0.18     \\
      & MBPP-Hard       & 47.75 & 49.55 & 2.70 & 62.56 & 33.33 & 4.11  \\
      & MBPP-Hard-Guided  & - & - & - & 72.44     & 26.28     & 1.28     \\
      & Leetcode-Original & - & - & -  & 38.60 & 58.77 & 2.63    \\
      & Leetcode-New  & 40.59 & 58.82 & 0.59 & 48.47 & 49.08 & 2.45  \\
      & Leetcode-Guided  & - & - & - & 46.33     & 52.54     & 1.13     \\
    \bottomrule
  \end{tabularx}
  }
\end{table*}

%% file: sections/6-conclusion.tex
\section{Conclusion}
In this paper, we introduced InfoSynth, a novel framework to calculate the diversity and novelty of new benchmarks in an efficient and cost-effective manner. Using the ideas behind this framework, we propose a new coding problem generation pipeline that produces more challenging and robust problems from seed data than previous works. We hope that future work will leverage our ideas to create robust, novel, and diverse benchmarks.

\section{Limitations}
InfoSynth has several limitations, which we explicitly acknowledge.

In a small number of cases, LLMs may generate correct solutions but fail to strictly conform to the required output format (e.g., function signatures or structured test specifications). Our execution-based validation pipeline mitigates most of these cases by enforcing deterministic programmatic checks, though improving semantic parsing is a natural extension. Our evaluation also focuses on Python problems; however, InfoSynth is language-agnostic and we believe the framework can be extended to other languages with executable verification.

Additionally, a significant fraction of candidate problems fail self-verification, including some that are otherwise novel or non-trivial. We believe that this filtering is an intentional design choice that prioritizes correctness and evaluability over raw generation volume. Importantly, we observe that higher rejection rates correlate with increased final benchmark difficulty, suggesting the filtering step is not merely discarding noise but actively shaping a more challenging problem distribution. For example, 28.2\% of problems generated by OpenCodeInstruct have failing solutions/tests, and up to 60\% of problems generated by KodCode are filtered out. However, as shown in Table \ref{tab:infosynth-comparison-pass-rate}, these problems are substantially easier than those generated by InfoSynth. Future improvements could incorporate stronger verifier models or multi-model consensus to further reduce false rejections while preserving this difficulty signal.

Moreover, our novelty and diversity metrics are computed using k-NN estimators in a UMAP-projected embedding space. While UMAP does not preserve global geometry exactly, it is designed to maintain local neighborhood structure, which is the primary signal used by k-NN estimators. As a result, the metrics should be interpreted as stable surrogate measures for relative comparison rather than absolute Shannon entropy or KL-divergence in the original space. While we do not have formal theoretical guarantees, we observe empirically that rankings induced by these metrics remain stable across embedding choices and hyperparameter settings, supporting their practical utility.

Finally, because of its filtration step, InfoSynth is biased toward short-horizon reasoning problems. Despite this constraint, we find that the resulting tasks remain challenging for frontier models, indicating that short-horizon difficulty is still a meaningful evaluation axis. Extending the framework to promote longer horizon tasks is a promising direction for expanding beyond this regime.

\section{Ethical considerations}
This paper presents work whose goal is to advance the field of machine learning by enabling scalable, information-guided synthesis of novel benchmarks for LLMs, addressing data creation challenges. Potential societal consequences include improved, unbiased evaluations of LLM capabilities in reasoning and code generation, fostering reliable AI deployment in applications like software development and scientific discovery. This work does not involve human subjects, personally identifiable data, or sensitive attributes. All datasets used (MBPP, Leetcode, and Codeforces) are publicly available, and our generated benchmarks were produced through synthetic problem generation and automated verification. We have carefully ensured that no private or proprietary code was included. Potential risks include the misuse of generated benchmarks for unfair evaluation or dataset contamination in future model training; to mitigate this, we document our pipeline in detail and encourage responsible use. We used a few open-source datasets: MBPP \cite{mbpp} which uses a CC-BY-4.0 license, a Leetcode dataset by \cite{leetcode-seed} which uses Apache 2.0, a dataset by \cite{leetcode-kl-div} which uses an MIT license, the APPS dataset \cite{apps-dataset} which uses CC-BY-SA-3.0, a Codeforces dataset \cite{openr1_codeforces} which uses CC-BY-4.0, and HumanEval \cite{human-eval} which uses the MIT license. We have open-sourced all code and datasets for InfoSynth under the Apache 2.0 license. Our work has used the datasets above only for research purposes, and is fully compatible with their access conditions. We also acknowledge that there was some usage of AI assistants for this work; however, they were used only to make minor grammatical improvements to the manuscript.

%% file: sections/appendix.tex
\section{Sensitivity Analysis for \texorpdfstring{$k$}{k} and UMAP Parameters}
\label{app:sensitivity}
We present ablations for hyper-parameters used in our experiments. In general, our results generalize robustly across a wide range of hyper-parameters. Note that in the cases where changing a hyperparameter affects the novelty or diversity, it affects all datasets by roughly the same amount. Since our analysis looks at \textit{relative} differences between benchmarks, our conclusions are still robust.

\begin{figure*}[htbp]
\centering

\begin{subfigure}{0.67\textwidth}
    \centering
    \includegraphics[width=\textwidth]{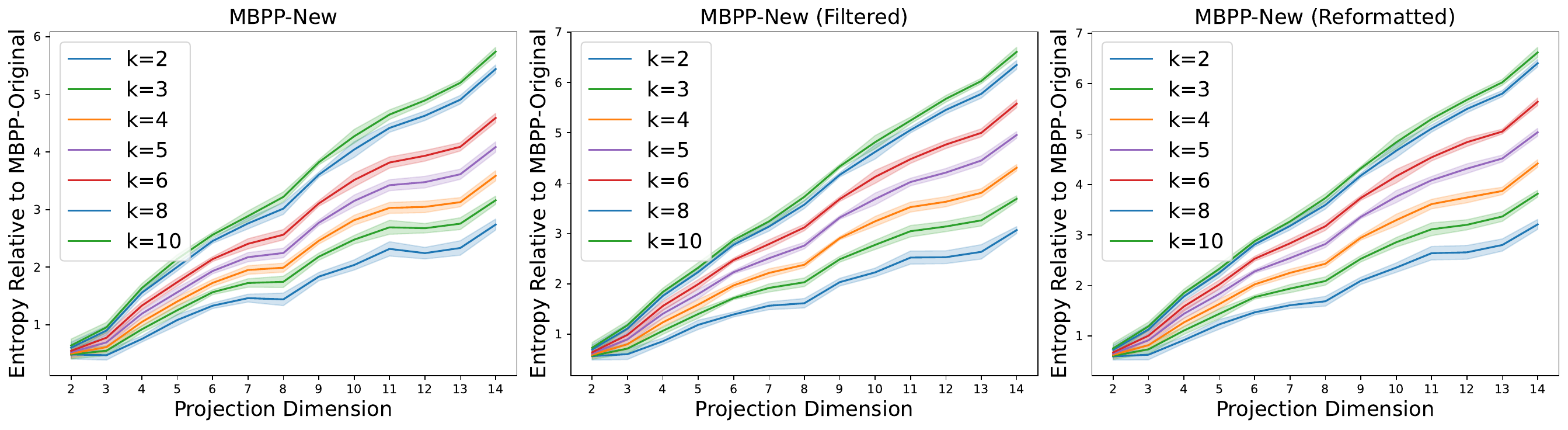}
    \caption{Diversity changes by a similar amount for all benchmarks as $k$ changes}
    \label{fig:sensitivity_k_entropy}
\end{subfigure}

\begin{subfigure}{0.67\textwidth}
    \centering
    \includegraphics[width=\textwidth]{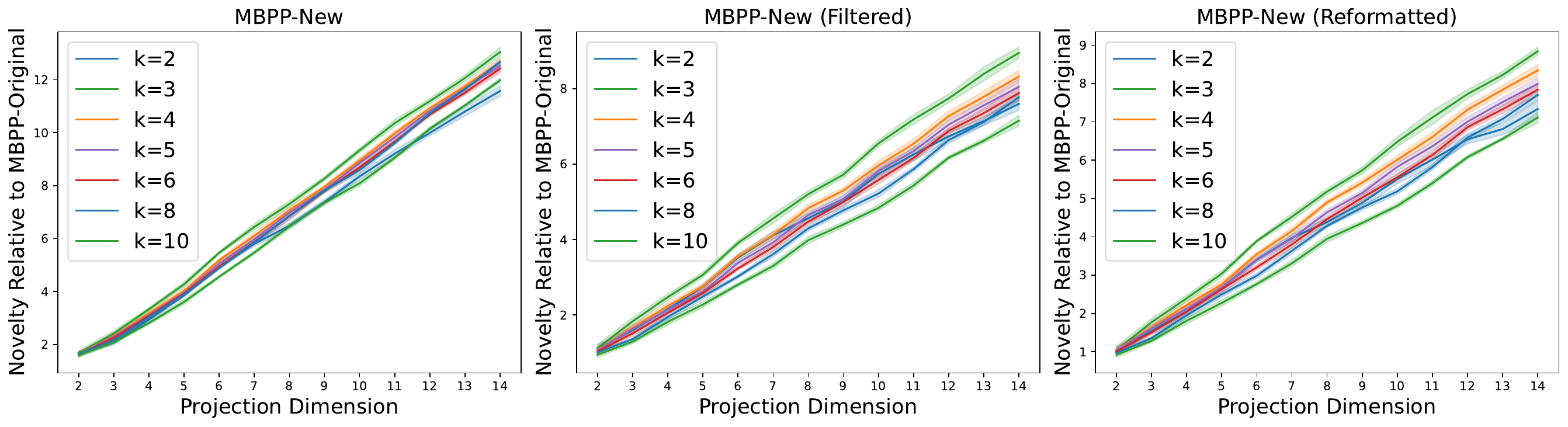}
    \caption{Novelty is almost invariant to fluctuations in $k$}
    \label{fig:sensitivity_k_kl}
\end{subfigure}

\begin{subfigure}{0.67\textwidth}
    \centering
    \includegraphics[width=\textwidth]{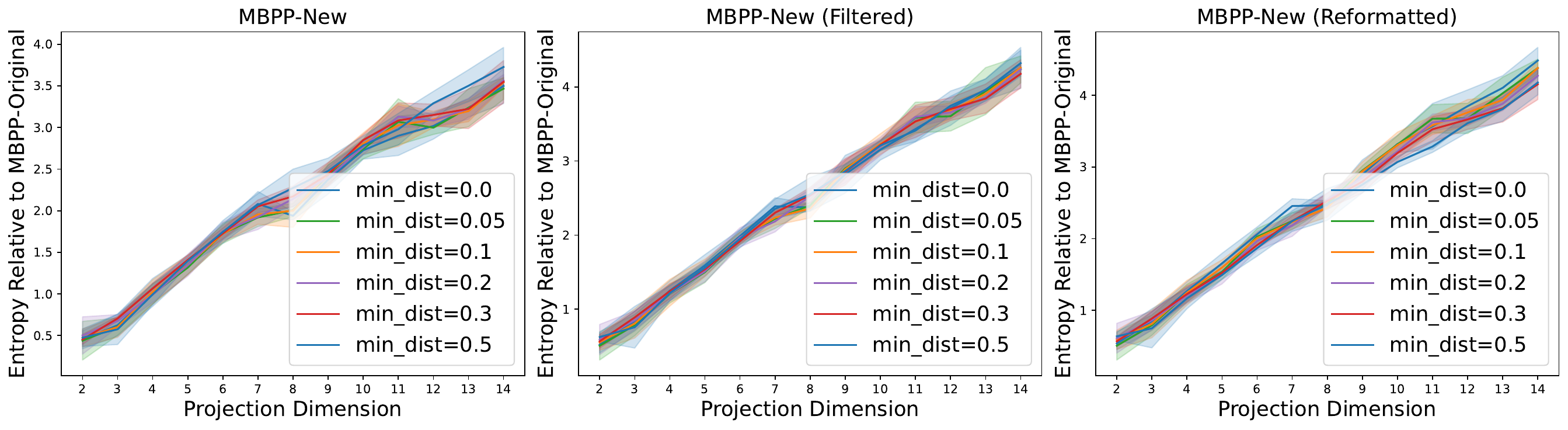}
    \caption{Diversity is almost invariant to fluctuations in n\_neighbors}
    \label{fig:sensitivity_neighbors_entropy}
\end{subfigure}

\begin{subfigure}{0.67\textwidth}
    \centering
    \includegraphics[width=\textwidth]{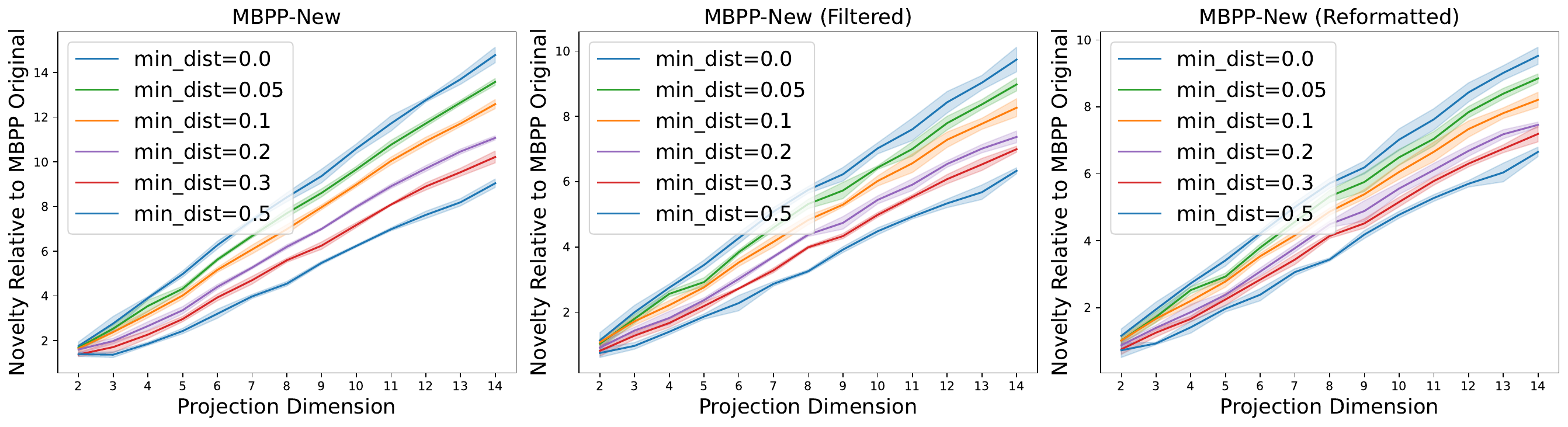}
    \caption{The novelty changes by a similar amount for all benchmarks as n\_neighbors changes}
    \label{fig:sensitivity_neighbors_kl}
\end{subfigure}

\begin{subfigure}{0.67\textwidth}
    \centering
    \includegraphics[width=\textwidth]{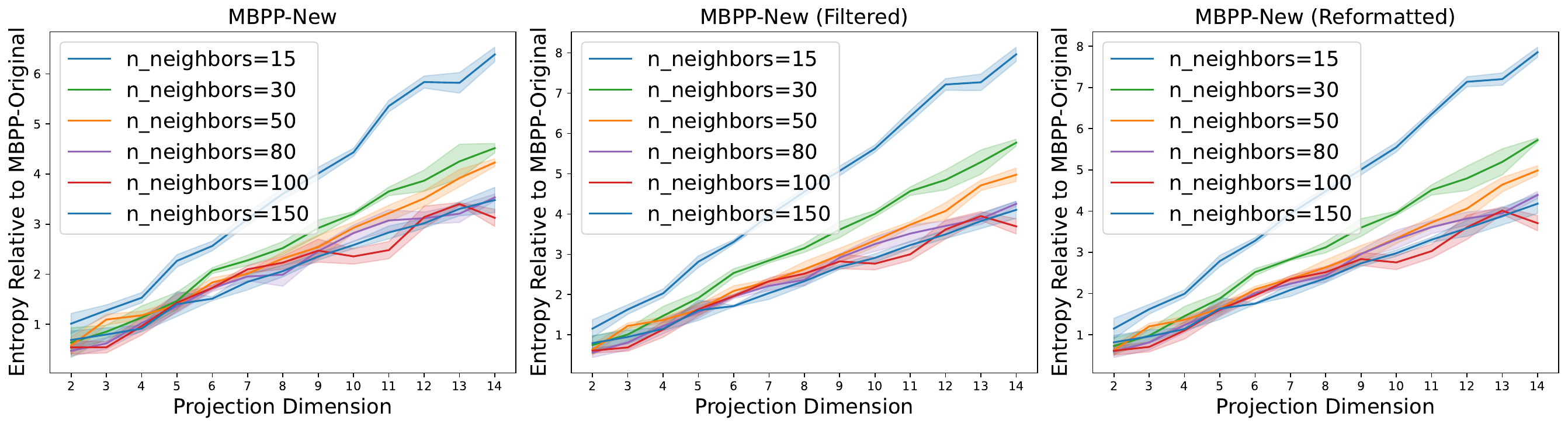}
    \caption{Diversity changes by a similar amount for all benchmarks as min\_dist changes}
    \label{fig:sensitivity_min_dist_entropy}
\end{subfigure}

\begin{subfigure}{0.67\textwidth}
    \centering
    \includegraphics[width=\textwidth]{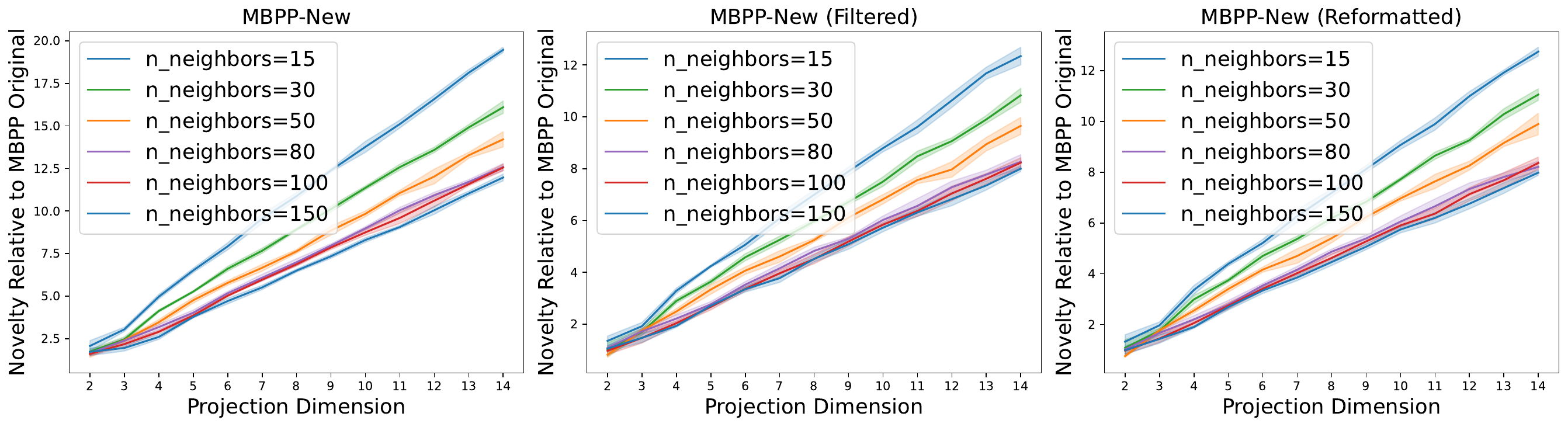}
    \caption{Novelty changes by a similar amount for all benchmarks as min\_dist changes}
    \label{fig:sensitivity_min_dist_kl}
\end{subfigure}

\caption{Sensitivity analysis across all hyperparameters.}
\label{fig:sensitivity_all}
\end{figure*}

Specifically, we ablate three hyperparameters in Figure \ref{fig:sensitivity_all}:
\begin{itemize}
    \item $k$: the number of k-NN neighbors considered in the estimators
    \item n\_neighbors: the number of neighbors used in the UMAP algorithm
    \item min\_dist: the minimum distance used in the UMAP algorithm
\end{itemize}

\section{Choice of Embedding Model}
\label{app:emb-model}
\begin{figure*}[htbp]
\centering
\begin{subfigure}{1\textwidth}
    \centering
    \includegraphics[width=\textwidth]{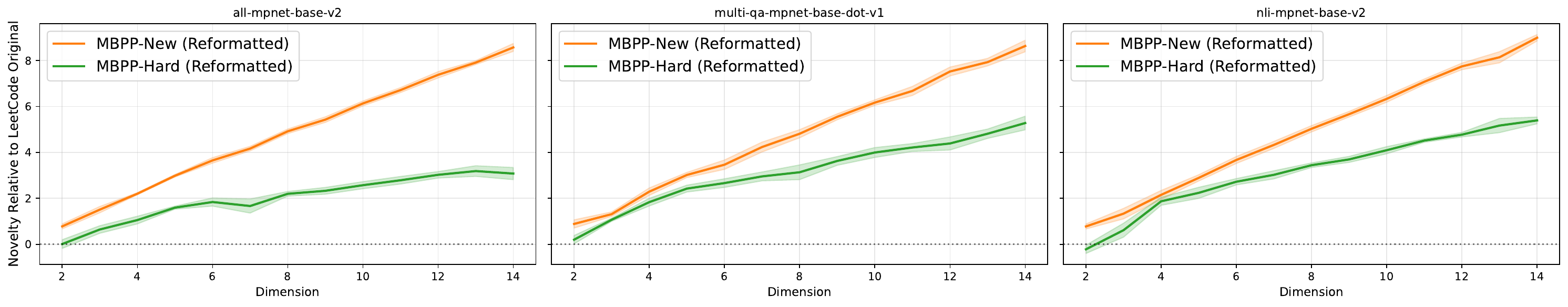}
    \caption{Novelty of datasets for various embedding models}
    \label{fig:kl_emb_invariance}
\end{subfigure}
\vspace{1em}  
\begin{subfigure}{1\textwidth}
    \centering
    \includegraphics[width=\textwidth]{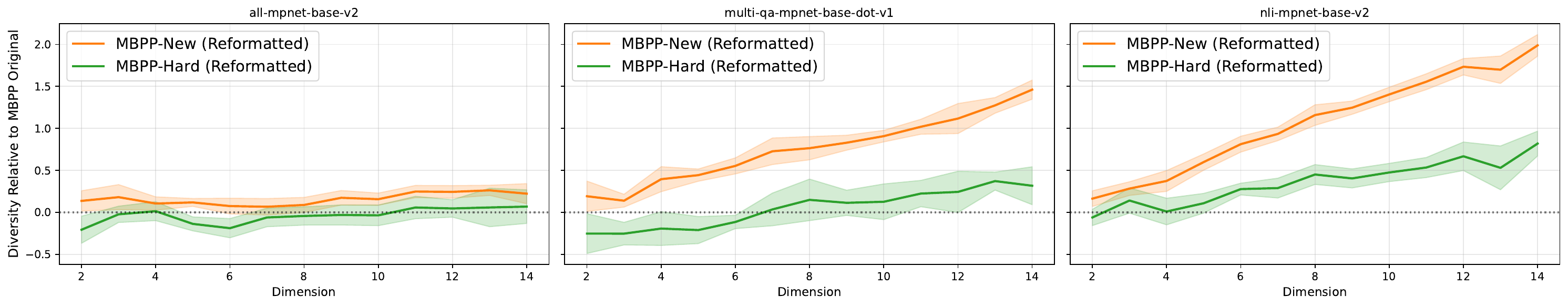}
    \caption{Diversity of datasets for various embedding models}
    \label{fig:entropy_emb_invariance}
\end{subfigure}
\caption{Embedding model invariance analysis for novelty and diversity metrics}
\label{fig:emb_invariance}
\end{figure*}

We test various embedding models on our datasets. Figures \ref{fig:kl_emb_invariance}, \ref{fig:entropy_emb_invariance} show that the relative novelty and diversity of datasets remains similar across embedding models despite some fluctuations in the magnitudes of those differences.

\section{Algorithm For Problem Generation}
\label{app:algo}
For MBPP-New we used $N = 1000$, $N_c = 10$, $B_s = 30$, $C = 2$, $B_c = 5$, $N_{it} = 5$. For MBPP-Hard we used $N = 500$, $N_c = 10$, $B_s = 15$, $C = 1$, $B_c = 4$, $N_{it} = 5$. For Leetcode-New we used $N = 1000$, $N_c = 10$, $B_s = 30$, $C = 2$, $B_c = 4$, $N_{it} = 5$.

For MBPP-Guided we used $N = 1000$, $N_c = 10$, $B_s = 30$, $C = 3$, $B_c = 5$, $N_{it} = 3$. For MBPP-Hard-Guided we used $N = 500$, $N_c = 10$, $B_s = 15$, $C = 3$, $B_c = 4$, $N_{it} = 3$. For Leetcode-Guided we used $N = 1000$, $N_c = 10$, $B_s = 30$, $C = 3$, $B_c = 4$, $N_{it} = 3$.

The algorithm is given in Algorithm \ref{app:algo_impl}

\begin{algorithm}[htbp]
\caption{Problem Generation with Evolutionary Strategies}
\label{app:algo_impl}
\small
\begin{algorithmic}
\STATE {\bfseries Input:} $N$: total number of problems; $N_c$: number of colonies; $B_s$: seed batch size; $C$: problems per crossover; $B_c$: crossover batch size; $N_{\text{it}}$: iterations
\STATE
\STATE {\bfseries procedure} \textsc{Generate}(seedData, numSamples)
\STATE Initialize problems $\gets \emptyset$
\FOR{colony $= 1$ {\bfseries to} $N_c$}
\STATE $N_s \gets N / N_c$
\STATE colonySeedData $\gets$ Random sample of size $B_s$ from seedData
\STATE problems $\gets$ problems $\cup$ \textsc{EvolveColony}(colonySeedData, $N_s$)
\STATE deduplicate(problems)
\ENDFOR
\STATE {\bfseries return} problems
\STATE
\STATE {\bfseries procedure} \textsc{EvolveColony}(seedData, $N_s$)
\STATE newProblems $\gets \emptyset$
\REPEAT
\STATE operation $\gets$ ``mutation'' w.p. 0.5, ``crossover'' w.p. 0.5
\IF{operation $==$ ``mutation''}
\STATE problem $\gets$ random sample from seedData
\STATE problems $\gets$ problems $\cup$ mutate(problem)
\ENDIF
\IF{operation $==$ ``crossover''}
\STATE batch $\gets$ random sample of size $B_c$ from seedData
\STATE problems $\gets$ problems $\cup$ crossover($C$, batch)
\ENDIF
\IF{$k$-farthest-neighbor-filtering enabled}
\STATE $U \gets$ newProblems $\cup$ seedData
\STATE problems $\gets$ select $K$ with least cosine similarity to $U$
\ENDIF
\STATE newProblems $\gets$ newProblems $\cup$ problems
\STATE deduplicate(newProblems)
\STATE seedData $\gets$ seedData $\cup$ problems
\UNTIL{$|$newProblems$| \geq N_s$}
\STATE {\bfseries return} newProblems
\STATE
\STATE {\bfseries procedure} \textsc{GenerateSolsTests}(problem)
\STATE GenerateTests(problem)
\STATE GenerateSolutions(problem)
\FOR{$i = 1$ {\bfseries to} $N_{\text{it}}$}
\STATE Run tests against solutions
\STATE Feed output to LLM to modify tests and solution
\ENDFOR
\end{algorithmic}
\end{algorithm}

\section{Code-Feedback Results}
\label{app:code-feedback}
\begin{figure}[htbp]
\centering
\begin{minipage}{0.45\textwidth}
    \centering
    \includegraphics[width=\textwidth]{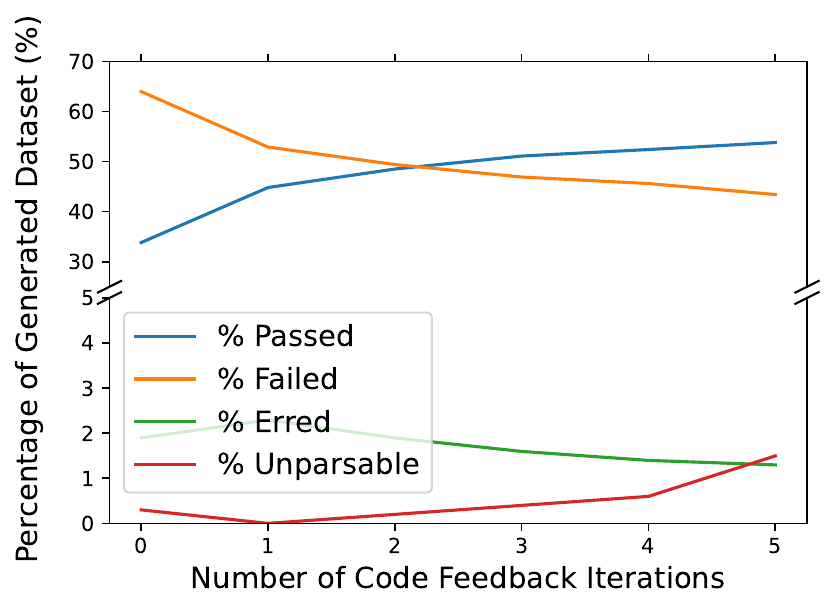}
    \captionsetup{width=0.8\textwidth}
    \caption{Self-verification for MBPP-New}
    \label{fig:self_verif_mbpp}
\end{minipage}
\begin{minipage}{0.45\textwidth}
    \centering
    \includegraphics[width=\textwidth]{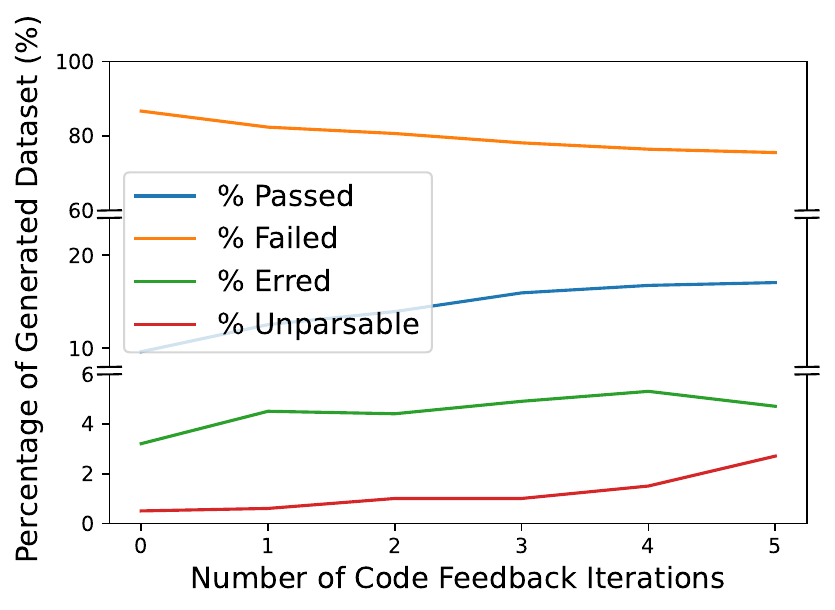}
    \captionsetup{width=0.8\textwidth}
    \caption{Self-verification for Leetcode-New}
    \label{fig:self_verif_lc}
\end{minipage}
\end{figure}

Figures \ref{fig:self_verif_mbpp}, \ref{fig:self_verif_lc} show how the proportion of problems that pass, fail, error, and are unparsable changes as a function of the number of code feedback iterations. The results shown are consistent across all datasets; in general, more than 3 feedback iterations provide minimal gains in pass rate.

\section{Analysis of Filtered-Out Problems}
\label{app:filtered-analysis}
In general, we do not observe strong systematic biases in the topics that survive the filtering stage. The retained dataset continues to cover the major categories present in the seed distribution, suggesting that the filtering process primarily removes invalid or inconsistent generations rather than collapsing the underlying problem space. However, we observe that certain classes of generated problems are disproportionately filtered out, and these cases also provide additional insight into model behavior under compositional generation and verification constraints.

The first such category consists of \textbf{deep crossover compositions}, where crossover is repeatedly applied to previously crossover-generated tasks. These produce highly compositional problems with tightly interacting constraints, which are difficult for the generator model to satisfy jointly while also producing consistent test cases. A second category involves \textbf{numerically sensitive problems}, such as recurrence-based or combinatorial computation tasks (e.g., computing the $n$th Delannoy number), where models often produce correct symbolic solutions but struggle to construct test cases with exact ground-truth values.

Across these filtered categories, we identify three recurring patterns that characterize model behavior:

\textbf{Generation Outpacing Verification:} To pass InfoSynth's pipeline, models must generate both code and test cases. In failed crossover problems, a frequent failure mode is that the LLM generates a correct solution but fails to generate tests with the correct input-output relationship. This reveals an asymmetry in capability: models can often reason abstractly about a class of solutions, but struggle to instantiate that reasoning into precise, verifiable test cases.

\textbf{Failure to Holistically Reason:} Crossover problems introduce multiple interacting constraints. We find that models frequently attempt to satisfy these sequentially rather than jointly. While this strategy can work for simpler tasks, it becomes brittle when constraints interact, as locally consistent decisions do not guarantee global correctness. This behavior is a key contributor to failure in deep crossover settings.

\textbf{Superficial Debugging Loops:} Although iterative execution feedback (Section \ref{iterative-code-feedback}) generally improves performance, it can stall on deeply nested crossover problems. The model often becomes trapped in a ``local minimum'' where it focuses on surface-level corrections such as syntax, formatting, or minor edge-case fixes. Even with multiple feedback iterations and access to its full interaction history, it fails to reliably revisit or reframe the underlying solution strategy. This suggests a limitation in stepping back to higher-level abstraction once an initial solution trajectory is committed.

In summary, the filtering process not only shapes the final dataset but also highlights systematic stress points in current LLMs. These observations suggest that the most frequently filtered problems are precisely those that require tightly coupled reasoning across solution generation and executable verification, providing a structured view into where current models remain fragile.

\section{Problem Generation Pipeline Prompts}
\label{app:prompts}
We note that our prompts share some similarity with those used by \citet{genetic-instruct}.
\subsection{Mutation Prompts}
\label{app:mutation}

\begin{lstlisting}[style=pythonstyle]   
% easy
Please decrease the difficulty of the given programming test question a bit. 
The new problem should be conceptually similar to the given question, 
but should not simply paraphrase it. Do not provide any hints, solutions 
or outputs. Only one new instruction is allowed.
Original Question: {instruction}
New Question:

% medium
Please create a new programming problem of the same difficulty as the
given programming test question. The new problem should be conceptually
similar to the given question, but should not simply paraphrase it. 
Do not provide any hints, solutions or outputs. Only one new instruction is allowed.
Original Question: {instruction}
New Question:

% hard
Please increase the difficulty of the given programming test question a bit.
Do not provide any hints, solutions or outputs. Only one new instruction is allowed.
Original Question: {instruction}
New Question:
\end{lstlisting}

\subsection{Crossover Prompt}
\label{app:crossover}

\begin{lstlisting}[style=pythonstyle]
I will provide you with a set of coding questions. Please give me a new coding
question that combines core concepts from two or more of the given questions. 
Please ensure that the new question is novel and does not simply paraphrase 
any of the problems I am giving you. Do not include any extra information 
that would help a test-taker other than the problem statement itself.

Question 1:
{instruction 1}

Question 2:
{instruction 2}
...

New Question:
\end{lstlisting}

\subsection{Solution \& Test Generation Prompt}
\label{app:solution-tests}
We note that our prompts are similar to those used by \citet{kodcode}.
\begin{lstlisting}[style=pythonstyle]
You are an expert in Python coding.
## Task:
Please answer the question and generate unit tests to verify your answer.

## Output Format:
Your solution and unit tests should be presented in the format within the
specified sections below. Ensure your code is within code blocks. For the
tests, use pytest style by defining individual test functions (without
classes) and using assert statements. Your tests should be implementation
independent. Ensure that you include the <|Solution Begin|>, 
<|Solution End|>, <|Test Begin|>, and <|Test End|> tags as depicted. The
solution function must be named solution.

<|Solution Begin|>
{Solution Code in Python}
<|Solution End|>

<|Test Begin|>
{Unit Test Code in Python}
<|Test End|>

## Example
Below is an example output format implementing a simple a + b function.
<|Solution Begin|>
def add(a, b):
    '''Returns the sum of a and b.'''
    return a + b
<|Solution End|>

<|Test Begin|>
from solution import add
def test_add_positive_numbers():
    assert add(2, 3) == 5
def test_add_with_zero():
    assert add(0, 5) == 5
    assert add(5, 0) == 5
def test_add_negative_numbers():
    assert add(-1, -1) == -2
def test_add_mixed_sign_numbers():
    assert add(-1, 3) == 2
<|Test End|>

## Question:
{problem}
\end{lstlisting}

\subsection{Solution \& Test Generation with Iterative Feedback Prompt}
\label{app:solution-tests-feedback}

\begin{lstlisting}[style=pythonstyle]
You are an expert in Python coding.
## Task:
Please answer the question and generate unit tests to verify your answer. The
entire chat history of your previous attempts to generate questions and unit
tests is presented below in the "Chat History" section, along with the output
of running your solution against your tests in a code execution environment.
Please modify only your tests and/or solution to be more correct.

## Output Format:
<Same as "Output Format" section above>

## Chat History:
Attempt 1 Solution:
{attempt 1 solution}

Attempt 1 Code Execution Output:
{attempt 1 code output}

Attempt 2 Solution:
{Attempt 2 solution}

Attempt 2 Code Execution Output:
{attempt 2 code output}
...

## Question:
{problem}
\end{lstlisting}

\subsection{Postprocessing Prompts}
\label{app:postprocess-prompt}
\begin{lstlisting}[style=pythonstyle]
You are an expert in Python coding. Here is a coding problem with associated test cases. Please rephrase the question so it describes what the user should output for edge-cases without changing the essence of the problem. Add as little information as possible, only describing what the user should output for edge-cases that cannot be inferred from the problem description. Do not include anything except for the rewritten problem in your response and do not include the test cases.
Question:
{question}
Tests:
{tests}
\end{lstlisting}

\section{Example of Post-processed Problems}
\label{app:post-processed}
\textbf{MBPP-New Dataset Original Problem:}
\begin{lstlisting}[style=pythonstyle]
Write a function that filters a list of usernames stored in a dictionary, returning only those associated with students who fall within a specified age range.
\end{lstlisting}
\noindent
\textbf{Post-Processed Version:}
\begin{lstlisting}[style=pythonstyle]
Write a function that filters a list of usernames stored in a dictionary, returning only those associated with students who fall within a specified age range. Ensure that the function returns an empty list when there are no students or when the input dictionary is empty.
\end{lstlisting}
\noindent
\textbf{Leetcode-New Dataset Original Problem:}
\begin{lstlisting}[style=pythonstyle]
Alice and Bob are engaged in a strategic game on an infinite 2D plane with n points provided by their coordinates in two integer arrays, xCoord and yCoord. They take turns, starting with Alice, attacking a point on the plane to capture it, with the condition that once a point is attacked, it is removed permanently from the game, and they have to remove exactly 1 point per turn.

The winner is the player who either removes the point that leaves no rectangle capable of being formed using the remaining points on their turn or can force the scenario by optimal play such that the opponent has no such move left on their subsequent turns.

Given the arrays xCoord and yCoord, along with knowledge of optimal strategies for both players, determine if Alice, who starts the game, can always guarantee a win. Return true if Alice has a winning strategy, or false if Bob can always force a win even with Alice starting first.
\end{lstlisting}
\noindent
\textbf{Post-Processed Version:}
\begin{lstlisting}[style=pythonstyle]
Alice and Bob are playing a strategic game on an infinite 2D plane with n points defined by their coordinates in two integer arrays, xCoord and yCoord. They alternately take turns, with Alice starting first, to attack and permanently remove exactly one point at a time. The objective is for a player to leave no possibility of forming a rectangle using any four of the remaining points. A player wins if they achieve this or if they can force a scenario where the opponent has no such moves left. Given the arrays xCoord and yCoord, determine if Alice has a guaranteed winning strategy. Return true if Alice can always win, and false if Bob can always force a win or if there are no points to start with.
\end{lstlisting}

\section{Examples of Generated Problems}
\label{app:problem-example}
We note that for conciseness, the examples shown in this section are before the postprocessing step. In this section, we illuminate the different mechanisms through which mutation and crossover generate interesting problems.
\subsection{Mutation Operation: Adding or Removing Constraints}
This example shows how mutation creates three variants of the question by adding requirements or removing constraints from the problem. We can see that all three problems require similar conceptual understanding, but the harder ones simply require more code or bookkeeping. \newline
\noindent
\begin{lstlisting}[style=pythonstyle]
Original Question:
Write a function to filter the height and width of students which are stored in a dictionary.
\end{lstlisting}

\begin{lstlisting}[style=pythonstyle]
Easy Mutation:
Write a function that filters a list of usernames stored in a dictionary, returning only those associated with students who fall within a specified age range. Ensure that the function returns an empty list when there are no students or when the input dictionary is empty.
\end{lstlisting}

\begin{lstlisting}[style=pythonstyle]
Medium Mutation:
Create a function that filters user profiles based on a dictionary. The function should return a list of user IDs for profiles where the age is within a specified range (inclusive) and the profile only contains lowercase alphabetic characters. If there are no valid profiles that match the criteria, the function should return an empty list.
\end{lstlisting}

\begin{lstlisting}[style=pythonstyle]
Hard Mutation:
Create a function to filter user passwords from a dictionary, returning only those that are valid for students whose dimensions (height, width, and weight) are within a given range. Each valid password must include at least one uppercase letter, one lowercase letter, one digit, and one special character. Ensure the solution appropriately handles and returns an empty dictionary for cases where no users are provided or all entries are invalid due to dimension or password criteria.
\end{lstlisting}

\subsection{Mutation Operation Example: Creative Modification}
This example shows how mutation creates three variants of the question by creatively modifying the central idea of the problem itself. This differs from the previous example; in this case, the harder questions require a fundamental understanding of new topics. \newline
\noindent
\begin{lstlisting}[style=pythonstyle]
Original Question:
Write a function to find the perimeter of a rectangle.
\end{lstlisting}

\begin{lstlisting}[style=pythonstyle]
Easy Mutation:
Write a function to find the area of a rectangle.
\end{lstlisting}

\begin{lstlisting}[style=pythonstyle]
Medium Mutation:
Write a function to calculate the area of a trapezoid given its base lengths and height.
\end{lstlisting}

\begin{lstlisting}[style=pythonstyle]
Hard Mutation:
Write a function to find the area of a rhombus given its diagonals and verify if the rhombus is also a square by using its side lengths.
\end{lstlisting}

\subsection{Crossover Operation: Combining Concepts}
This example shows how crossover creates an interesting, novel question by combining two unrelated concepts. A key difference between crossover and mutation is that crossover does not introduce any new concepts or content into the generated problem as it draws from existing ones. This demonstrates the necessity of combining mutation and crossover into one pipeline; mutation introduces new concepts into the dataset while crossover takes existing concepts and uses them to create richer problems.\newline

\noindent
\textbf{Seed Problems:}
\begin{lstlisting}[style=pythonstyle]
Question 1:
There is a 50 x 50 chessboard with one knight and some pawns on it. You are given two integers kx and ky where (kx, ky) denotes the position of the knight, and a 2D array positions where positions[i] = [xi, yi] denotes the position of the pawns on the chessboard.
Alice and Bob play a turn-based game, where Alice goes first. In each player's turn:

The player selects a pawn that still exists on the board and captures it with the knight in the fewest possible moves. Note that the player can select any pawn, it might not be one that can be captured in the least number of moves.
In the process of capturing the selected pawn, the knight may pass other pawns without capturing them. Only the selected pawn can be captured in this turn.

Alice is trying to maximize the sum of the number of moves made by both players until there are no more pawns on the board, whereas Bob tries to minimize them.
Return the maximum total number of moves made during the game that Alice can achieve, assuming both players play optimally.
Note that in one move, a chess knight has eight possible positions it can move to, as illustrated below. Each move is two cells in a cardinal direction, then one cell in an orthogonal direction.
\end{lstlisting}

\begin{lstlisting}[style=pythonstyle]
Question 2:
You are given an array points where points[i] = [xi, yi] represents the coordinates of a point on an infinite plane.
Your task is to find the maximum area of a rectangle that:

Can be formed using four of these points as its corners.
Does not contain any other point inside or on its border.
Has its edges parallel to the axes.

Return the maximum area that you can obtain or -1 if no such rectangle is possible.
\end{lstlisting}

\noindent
\textbf{Crossover Generated Question:}
\begin{lstlisting}[style=pythonstyle]
Alice and Bob are playing a strategic game on an infinite 2D plane with n points defined by their coordinates in two integer arrays, xCoord and yCoord. They alternately take turns, with Alice starting first, to attack and permanently remove exactly one point at a time. The objective is for a player to leave no possibility of forming a rectangle using any four of the remaining points. A player wins if they achieve this or if they can force a scenario where the opponent has no such moves left. Given the arrays xCoord and yCoord, determine if Alice has a guaranteed winning strategy. Return true if Alice can always win, and false if Bob can always force a win or if there are no points to start with.
\end{lstlisting}

\subsection{Crossover Operation: Multistep Reasoning}
This example shows how crossover sometimes creates problems that require multistep reasoning. The new problem effectively requires a test-taker to solve both the seed problems. \newline

\noindent
\textbf{Seed Problems:}
\begin{lstlisting}[style=pythonstyle]
Question 1:
Write a python function to find the sum of fourth power of first n odd natural numbers.
\end{lstlisting}

\begin{lstlisting}[style=pythonstyle]
Question 2:
Write a python function to find the sum of xor of all pairs of numbers in the given array.
\end{lstlisting}

\noindent
\textbf{Crossover Generated Question:}
\begin{lstlisting}[style=pythonstyle]
Write a python function to compute the combined metric of two properties for a given array of integers: 

1. The sum of the fourth power of the first 'k' odd numbers from the array.
2. The sum of the bitwise XOR of all pairs of numbers formed by selecting one element from the first 'k' even numbers and one element from the last 'k' elements in the array.

The function should take two arguments - the array of integers and an integer 'k'. Ensure that 'k' is less than or equal to half the length of the array and that the array contains a sufficient number of odd and even numbers.
\end{lstlisting}


\section{Example of Chain-of-Thought Thinking}
\label{app:cot}
This example shows how the model is able to reason about why its solution and tests are erroring due to receiving feedback from the code execution environment. \newline

\noindent
\textbf{Question:} Write a function that takes a list of strings and returns the second shortest string that contains all vowels, sorted alphabetically. If no such string exists, return an empty string.

\noindent
\textbf{Initial Solution:}
\begin{lstlisting}[style=pythonstyle]
def solution(strings):
    """
    Returns the second shortest string containing all vowels, sorted alphabetically.
    """
    all_vowels = set('aeiou')
    
    # Filter strings that contain all vowels
    filtered_strings = [s for s in strings if all_vowels <= set(s)]
    
    # Sort them by length first and then alphabetically
    filtered_strings.sort(key=lambda x: (len(x), x))
    
    # Return the second shortest string or empty string if it doesn't exist
    return filtered_strings[1] if len(filtered_strings) > 1 else ''
{Tests Omitted for Conciseness}
\end{lstlisting}

\noindent
\textbf{Solution after First Feedback iteration:}
\begin{lstlisting}[style=pythonstyle]
Let's revise the test cases to ensure they correctly test the function's behavior and check the solution implementation and its handling of edge cases. First, we'll make sure the solution properly handles strings with varying cases (uppercase and lowercase) and that it verifies the full vowel set in a case-insensitive manner. Additionally, let's adjust the tests to cover more scenarios and better reflect the conditions described in the question. Here's the revised version of the solution and unit tests:

{Solutions Omitted for Conciseness}
{Tests Omitted for Conciseness}
\end{lstlisting}

\noindent
\textbf{Solution after Second Feedback iteration:}
\begin{lstlisting}[style=pythonstyle]
We need to ensure that our solution and tests align correctly with the task. Let's refine both the solution and the tests so that they pass correctly according to the given problem statement. The solution needs to ensure that we filter strings containing all vowels in any case, sort them, and return the second shortest. The test cases should reflect this logic.

Here's the refined solution and tests:

{Solutions Omitted for Conciseness}
{Tests Omitted for Conciseness}
\end{lstlisting}

\section{Topic-Labeling Prompt}
\label{topic-labeling}
\begin{lstlisting}[style=pythonstyle]
You are a precise tagger for coding interview problems.

Given a problem statement and solution, choose up to THREE topics from the bank below that best describe the core techniques/data structures needed to solve it. Return JSON ONLY, no extra text, as:
{"topics": ["Topic1", "Topic2"]}

Rules:
- Use EXACT spellings from the bank (case/spacing must match).
- Prefer the most specific tag available (e.g., "Binary Tree" over "Tree", "Shortest Path" over "Graph" when appropriate).
- If the solution critically relies on a data structure (e.g., "Heap (Priority Queue)"), include it.
- If multiple techniques are essential (e.g., DP + Bitmask), include both.
- Do NOT exceed 3 topics; order them by importance.
- If nothing fits, choose the closest general tag (e.g., "Graph", "Array", "Math"), never invent tags.

Topic Bank (allowed values only):
Array; String; Hash Table; Dynamic Programming; Math; Sorting; Greedy; Depth-First Search; Binary Search; Database; Matrix; Tree; Breadth-First Search; Bit Manipulation; Two Pointers; Prefix Sum; Heap (Priority Queue); Simulation; Binary Tree; Graph; Stack; Counting; Sliding Window; Design; Enumeration; Backtracking; Union Find; Linked List; Number Theory; Ordered Set; Monotonic Stack; Segment Tree; Trie; Combinatorics; Bitmask; Divide and Conquer; Queue; Recursion; Geometry; Binary Indexed Tree; Memoization; Hash Function; Binary Search Tree; Shortest Path; String Matching; Topological Sort; Rolling Hash; Game Theory; Interactive; Data Stream; Monotonic Queue; Brainteaser; Doubly-Linked List; Randomized; Merge Sort; Counting Sort; Iterator; Concurrency; Probability and Statistics; Quickselect; Suffix Array; Line Sweep; Minimum Spanning Tree; Bucket Sort; Shell; Reservoir Sampling; Strongly Connected Component; Eulerian Circuit; Radix Sort; Rejection Sampling; Biconnected Component

Input:
[Problem]
{problem}

[Solution]
{solution}

Output:
JSON with key "topics" and UP TO 3 strings from the bank. No prose, no explanations.
"""

\end{lstlisting}